\documentclass[fleqn,10pt]{article}
\usepackage[utf8]{inputenc}
\usepackage[T1]{fontenc}

\usepackage{graphicx}
\usepackage{subfig}
\usepackage{multicol}
\usepackage{float} 
\usepackage{booktabs}
\usepackage{multirow} 
\usepackage{url}
\usepackage{amsmath}

\usepackage{caption} 

\usepackage[style=numeric,backend=biber]{biblatex}
\usepackage{siunitx}  
\DeclareSIUnit\pixel{px}

\usepackage{authblk} 

\title{LEyes: A Lightweight Framework for Deep Learning-Based Eye Tracking using Synthetic Eye Images}

\author[1,*]{Sean Anthony Byrne}
\author[2]{Virmarie Maquiling}
\author[3]{Marcus Nystr{\"o}m}
\author[2]{Enkelejda Kasneci}
\author[3,4,*]{Diederick C. Niehorster}
\affil[1]{MoMiLab, IMT School for Advanced Studies Lucca, Lucca, Italy}
\affil[2]{Human-Centered Technologies for Learning, Technical University of Munich, Munich, Germany}
\affil[3]{Lund University Humanities Lab, Lund University, Lund, Sweden}
\affil[4]{Department of Psychology, Lund University, Lund, Sweden}
\affil[*]{sean.byrne@imtlucca.it, diederick\_c.niehorster@humlab.lu.se}

\begin{document}
\maketitle

\begin{abstract}
Deep learning has bolstered gaze estimation techniques, but real-world deployment has been impeded by inadequate training datasets. This problem is exacerbated by both hardware-induced variations in eye images and inherent biological differences across the recorded participants, leading to both feature and pixel-level variance that hinders the generalizability of models trained on specific datasets. While synthetic datasets can be a solution, their creation is both time and resource-intensive. To address this problem, we present a framework called Light Eyes or "LEyes" which, unlike conventional photorealistic methods, only models key image features required for video-based eye tracking using simple light distributions. LEyes facilitates easy configuration for training neural networks across diverse gaze-estimation tasks. We demonstrate that models trained using LEyes are consistently on-par or outperform other state-of-the-art algorithms in terms of pupil and CR localization across well-known datasets. In addition, a LEyes trained model outperforms the industry standard eye tracker using significantly more cost-effective hardware. Going forward, we are confident that LEyes will revolutionize synthetic data generation for gaze estimation models, and lead to significant improvements of the next generation video-based eye trackers.
\end{abstract}

\section*{Main}

Gaze estimation refers to the computational techniques employed to ascertain an individual's point of visual focus. Commonly, algorithms in this field use eye images as inputs and yield an inferred gaze point or gaze direction, typically represented as x,y coordinates~\cite{9153754}. This field of research has recently witnessed a surge of new interest driven by technological advancements across various domains. Most notably, the widespread adoption of Virtual Reality (VR) headsets~\cite{garbin2020dataset, palmero2021openeds2020}, the integration of eye tracking technology into smartphones and tablets~\cite{valliappan2020accelerating, krafka2016eye}, and the continuous improvement of both wearable eye tracking devices~\cite{santini2019grip} and augmented reality systems~\cite{renner2017attention} have fueled this growing interest. Further, there has been a remarkable expansion of high-resolution eye tracking experiments recorded in controlled laboratory settings, where participants are often positioned with chin and forehead rests to enable precise eye movement to be captured. This research spans across diverse domains, such as healthcare~\cite{pierce2016eye}, economics~\cite{lahey2016power,byrne2023predicting, 10.1145/3591130}, neuroscience and cognitive science~\cite{hessels2019eye,niehorster2019searching}, and education~\cite{strohmaier2020eye,10.1145/3588015.3589197}, highlighting the extensive potential and broad applicability of eye tracking technology.

A prevalent gaze estimation technique involves recording videos of eye movements and monitoring the position of both the pupil (P) and any corneal reflections (CR) present in each frame. This process, known as P-CR eye tracking, together with a quick calibration procedure estimates where a person is looking and the movement of their eyes~\cite{kar2017review}. Beyond enhancing the precision of the device itself, a more accurate gaze estimation enables other less obvious benefits such as improved foveated rendering, optimizing GPU resources by rendering detailed areas while reducing peripheral resolution~\cite{kar2017review, walton2021beyond}. This lowers the computational load of the system and improves the visual experience for the user~\cite{kothari2022ellseg, palmero2021openeds2020}. Improved gaze estimation also facilitates natural interactions in virtual environments through realistic eye contact between avatars~\cite{yassien2020design}, assists users with mobility impairments~\cite{meena2017multimodal}, and serves as a key tool for technical training and evaluation in fields such as surgery~\cite{tien2014eye, chetwood2012collaborative}, dentistry~\cite{castner2020deep}, and aviation~\cite{niehorster2020towards}.

The deployment of deep learning algorithms has significantly enhanced the accuracy and robustness of gaze estimation techniques, as evidenced by multiple studies~\cite{fuhl2016pupilnet, fuhl2017pupilnet, fuhl2023pistol, maquiling2023virnet, Ellseg_gen, kim2019nvgaze,nair2020rit}. Deep learning algorithms address issues present in conventional algorithmic approaches, which are vulnerable to unpredictable factors like blinks or reflections in the recording~\cite{kothari2022ellseg}. Yet despite these benefits, the incorporation of deep learning algorithms continues to pose challenges, principally due to the complex task of gathering data for training the model~\cite{garbin2020dataset,palmero2021openeds2020}. This data procurement obstacle in gaze estimation can be detailed as follows:

\begin{figure*}
    \centering
    \includegraphics[width=.88\textwidth]{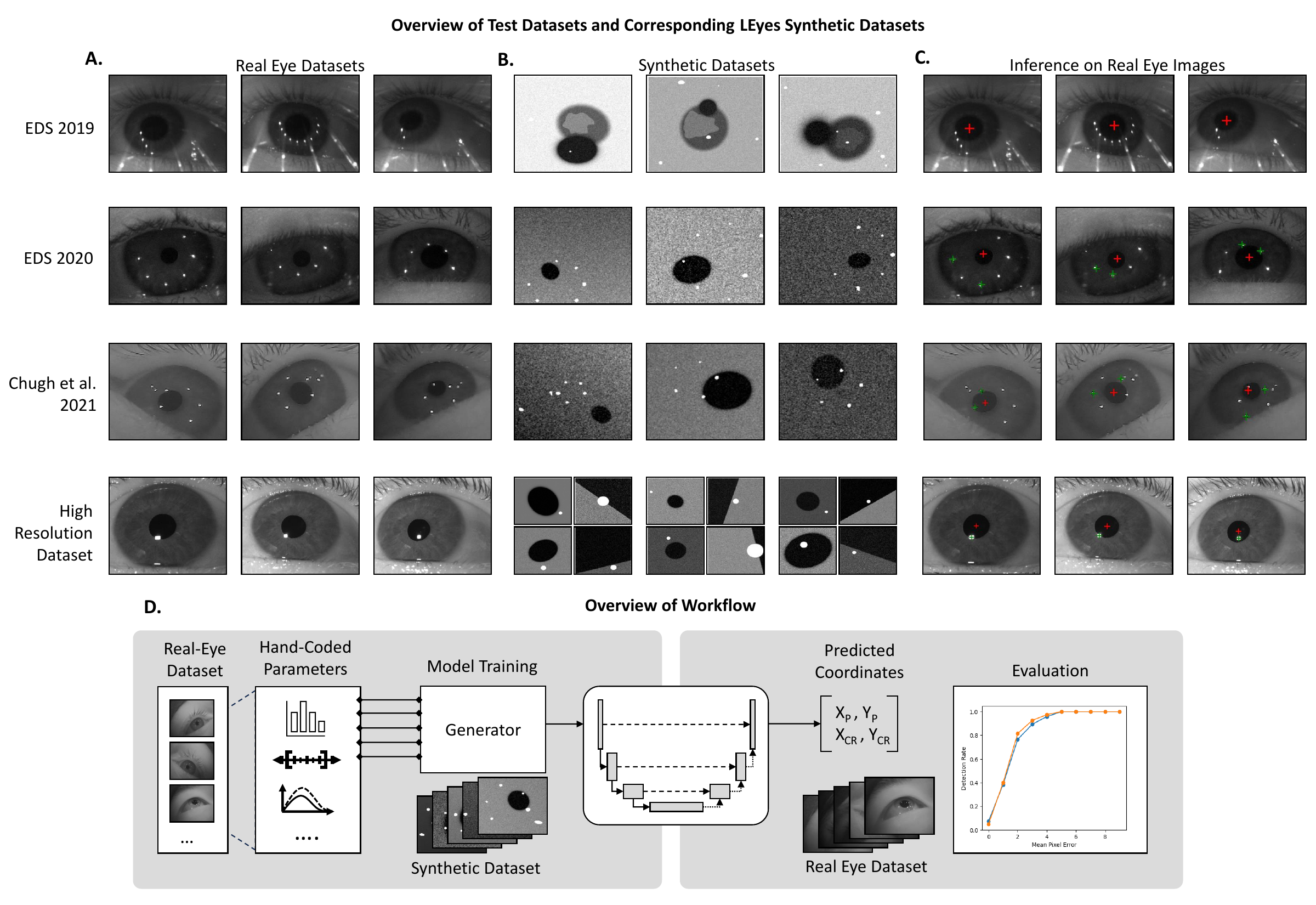}
    \caption{\textbf{A.} Images from the four datasets we used to test the LEyes framework. \textbf{B.} The LEyes synthetic training sets corresponding to the real eye datasets in A. These images are based on the light distributions of the real eye datasets. \textbf{C.} This shows the predictions of the LEyes trained model on the real eye images. 
    \textbf{D.} An overview of our approach: First, we establish a set of parameters based on the distributions of the collected data. These distributions pertain to pixel-level details like the iris and pupil intensity. Next, we employ a generator to efficiently produce new synthetic images from these parameters. The generated images are used to train a neural network which is then tested on real eye images recorded from the same device.}
    \label{fig:highreset}
    \vspace*{-0.25cm}

\end{figure*}

\begin{enumerate}

    \item ~\textbf{Data scarcity:} While data scarcity is a common issue across many deep learning domains \cite{bansal2022systematic}, this challenge is particularly acute in the field of eye-tracking research. Collecting a sufficient amount of training data for the development of deep learning models in this area demands significant time and resources \cite{byrne2023CRCNN,garbin2020dataset}.
    \item \textbf{Annotated datasets:} The second challenge involves the necessity for annotating segmented regions within eye images. This annotation is essential for creating labels for supervised learning algorithms that train deep learning models. It is a process that not only is time-consuming but also technically demanding, often requiring manual labeling by an experienced researcher~\cite{garbin2020dataset,palmero2021openeds2020}.
    \item \textbf{Differences in recorded eye images:} The third challenge stems from disparities in eye images found in the limited amount of publicly available datasets. Differences can occur not just across recording setups, but also from variation in eye image attributes like iris brightness, which lead to pixel level differences that contribute to sub-optimal network performance~\cite{nair2020rit}. This is a major issue as slight differences can have a substantial impact model performance and generalizability.
\end{enumerate}


A proposed solution to these challenges is the use of synthetic datasets which allows for the generation of vast amounts of annotated images~\cite{kim2019nvgaze, byrne2023CRCNN, maquiling2023virnet, nair2020rit}. Synthetic data has been used successfully to train deep neural networks in fields such as medical imaging~\cite{gao2023synthetic}, autonomous driving~\cite{osinski2020simulation}, and microscopy~\cite{helgadottir2019digital}. Typically in the field of gaze estimation, synthetic eye images creation methods aims for photorealism by employing a 3D model of the human eye and surrounding facial region to produce 2D images akin to those captured by eye-trackers, using render software or game engines such as Blender or Unity. The goal of such processes is to match the synthetic dataset's underlying distribution with the variability seen in real-world eye images~\cite{Wood_2015_ICCV,nair2020rit}. The photorealistic synthetic data approach, however, is not without limitations. One key challenge is the complexity of generating synthetic datasets that accurately emulate the distribution of real eye images. Additionally, concerns exist regarding the potential for achieving state-of-the-art outcomes when compared to models trained on genuine eye images. A study illustrated a decline in model accuracy by $1^\circ$ when comparing a model trained on photorealistic synthetic images to one trained on a subset of real eye images using a neural network~\cite{kim2019nvgaze,nair2020rit}. We hypothesize that numerous intricate features must be precisely constructed during synthetic dataset creation, and even minor deviations in design can significantly impact a model's inference capabilities during testing.



In this study, we adopt an innovative approach that departs from the traditional practice of creating photorealistic images, choosing instead to capitalize on the inherent simplicity of eye images. Rather than meticulously recreating every visual aspect, we focus on modeling the light distributions of the key features within an eye image required for eye tracking. We have found that LEyes images are not only easy to create, but are fast to generate. Most importantly, our approach based on LEyes provides more accurate results than other synthetic data methods over a range of different eye tracker setups.


\section*{Results}

\subsection*{Overview of the LEyes Framework}



Previous research~\cite{byrne2023CRCNN, nystrom2023amplitude, maquiling2023virnet} has shown that key features in eye images relevant for eye tracking can be effectively represented using 2D Gaussian distributions. Creating a synthetic dataset of eye images necessitates the accurate portrayal of such features, including the pupil, reflections, and pixel-level characteristics such as iris brightness~\cite{kim2019nvgaze, nair2020rit, Ellseg_gen}, which can be affected by specific lighting conditions altering dimensions and luminosity of features located in the image such as the iris or pupil. Emulating essential hardware attributes, such as lighting conditions and camera parameters, is vital for replicating real-world situations~\cite{kim2019nvgaze, nair2020rit, Ellseg_gen}. LEyes shows that by generating abstract images using 2D Gaussian distributions that contain the relevant features for an eye tracker, one can effectively capture both eye features and camera attributes for neural network training. The approach is outlined as follows:

First, to model key features such as the pupil, iris, and CRs, luminance attributes are derived by calculating the distributions of recorded data on a given device setup. To ensure generalizability of the model for a wide range of participants we use a larger parameter range than is derived from the distributions. Subsequently, these parameter ranges calculated from the distributions are utilized to craft images by layering and combining the parameter inputs through mathematical operations, achieving simple but realistic portrayals of eye features and noise within the created image. The images are then scaled and discretized to align with standard 8-bit camera output. Refer to the Methods section for a complete description of the process of generating LEyes images along with full descriptions of each model architecture used in the paper. 

To turn the feature parameters into images to be used to train the deep learning models we utilize the generator function from the DeepTrack 2.1 package~\cite{midtvedt2021quantitative}. The use of a generator combined with the relatively simple images created from 2D Gaussian distributions enables swift creation of customized synthetic images at reduced computational cost compared to photorealistic models. For example, the NVGaze dataset required 30 seconds to create each image; to create the entire dataset, would take approximately 3.8-years on a single GPU. In practice, this was reduced to a week as the researchers had access to a supercomputer~\cite{kim2019nvgaze}. Our models require no special computational resources and can be trained on platforms such as Google Colab, making them accessible to a wider group of researchers. The generator function also keeps track of both the image and corresponding label during training, which allows the generator to discard images after one pass to prevent over-training. Importantly, no images need to be pre-generated and occupy disk-space when a generator function is used~\cite{helgadottir2019digital}.

\subsection*{Pupil Localization}
We begin our analysis by considering the performance of the LEyes framework in a pupil center localization task in a VR setting, a common task for video based eye trackers~\cite{kim2019nvgaze}. To test our model we selected the widely used 2019 EDS challenge dataset (OpenEDS 2019)~\cite{garbin2020dataset}. We chose this dataset to run a comparative analysis as it has been used extensively to assess the accuracy of other methods in gaze estimation tasks~\cite{nair2020rit, chaudhary2022temporal, kim2019eye, Ellseg_gen, kothari2022ellseg}. 

OpenEDS 2019~\cite{garbin2020dataset} was collected using a VR head-mounted display equipped with dual eye-facing cameras, capturing images at 200 Hz under controlled lighting conditions. The dataset encompasses eye-region video footage from 152 participants for a total of 12,759 images featuring pixel-level annotations derived from human-annotated key points of the iris, pupil, and sclera. For a complete description of the data, refer to the original paper~\cite{garbin2020dataset}.

Various deep learning architectures have been proposed for eye segmentation tasks and LEyes simulations are model agnostic, yet, in light of their prevalent use and proven efficacy in eye tracking tasks~\cite{chugh2021detection,wang2023eye}, we chose to train a U-Net model with a ResNet-34 backbone. The model takes a grayscale eye image as its input and outputs a probability map indicating the location of the pupil in the image. To determine the center of the pupil we threshold this mask and employ a center of mass algorithm on the pupil region in the resulting binary image.

We compare our results with other state-of-the-art models and frameworks, including Pistol~\cite{fuhl2023pistol}, PuRe~\cite{santini2018pure}, the EllSeg framework~\cite{kothari2022ellseg,Ellseg_gen}, and DeepVog~\cite{yiu2019deepvog}. These models employ a variety of methods, ranging from conventional ellipse fitting to deep learning architectures. Note that we stress the difference between model and framework where a ``model'' is a specific representation trained to make predictions, while a ``framework'' is a set of tools and libraries used to develop, train, and deploy such models.

\begin{figure*}%
\centering
\subfloat[]{\includegraphics[height=0.31\textwidth]{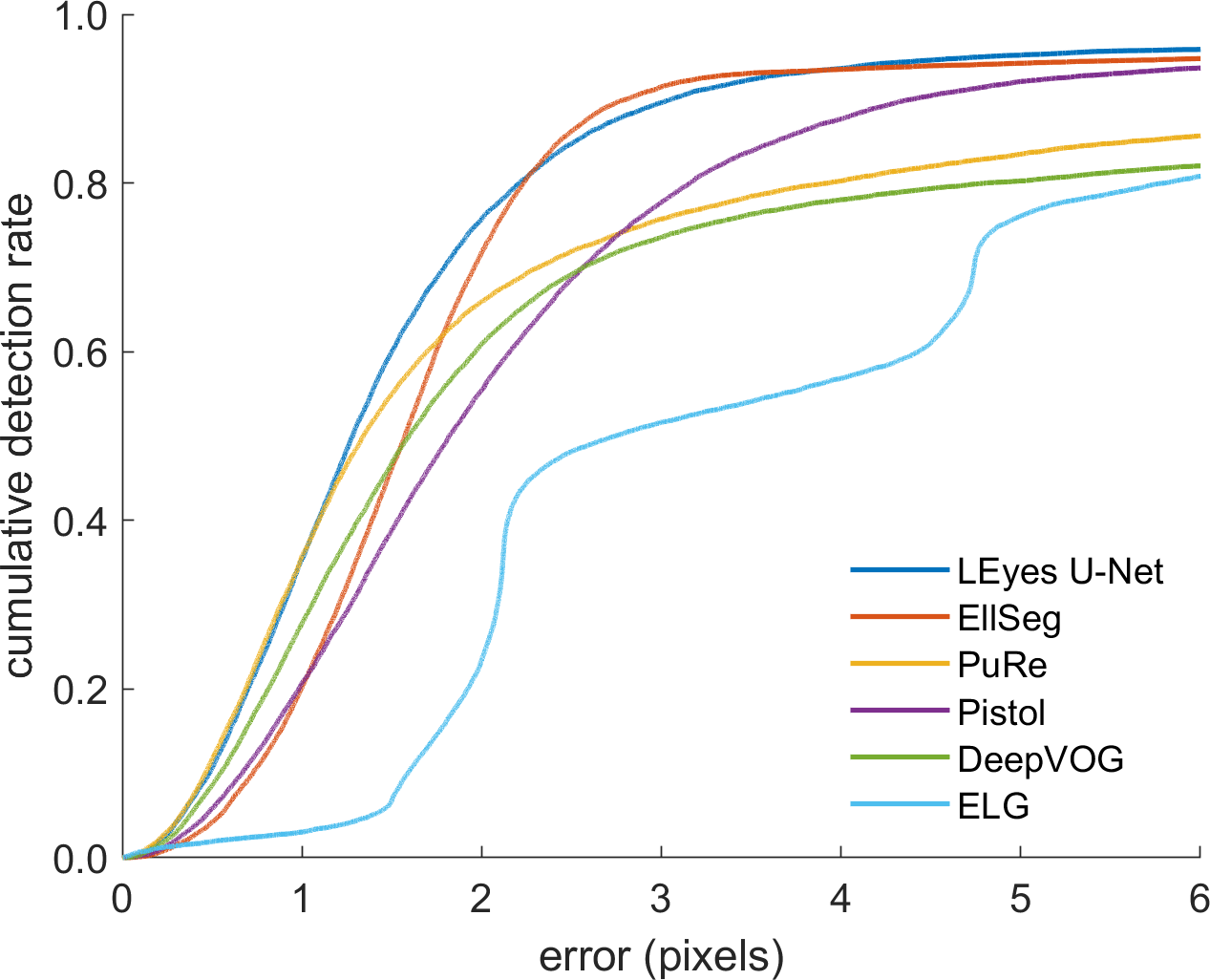}}
\hspace{1em}
\subfloat[]{\includegraphics[height=0.31\textwidth]{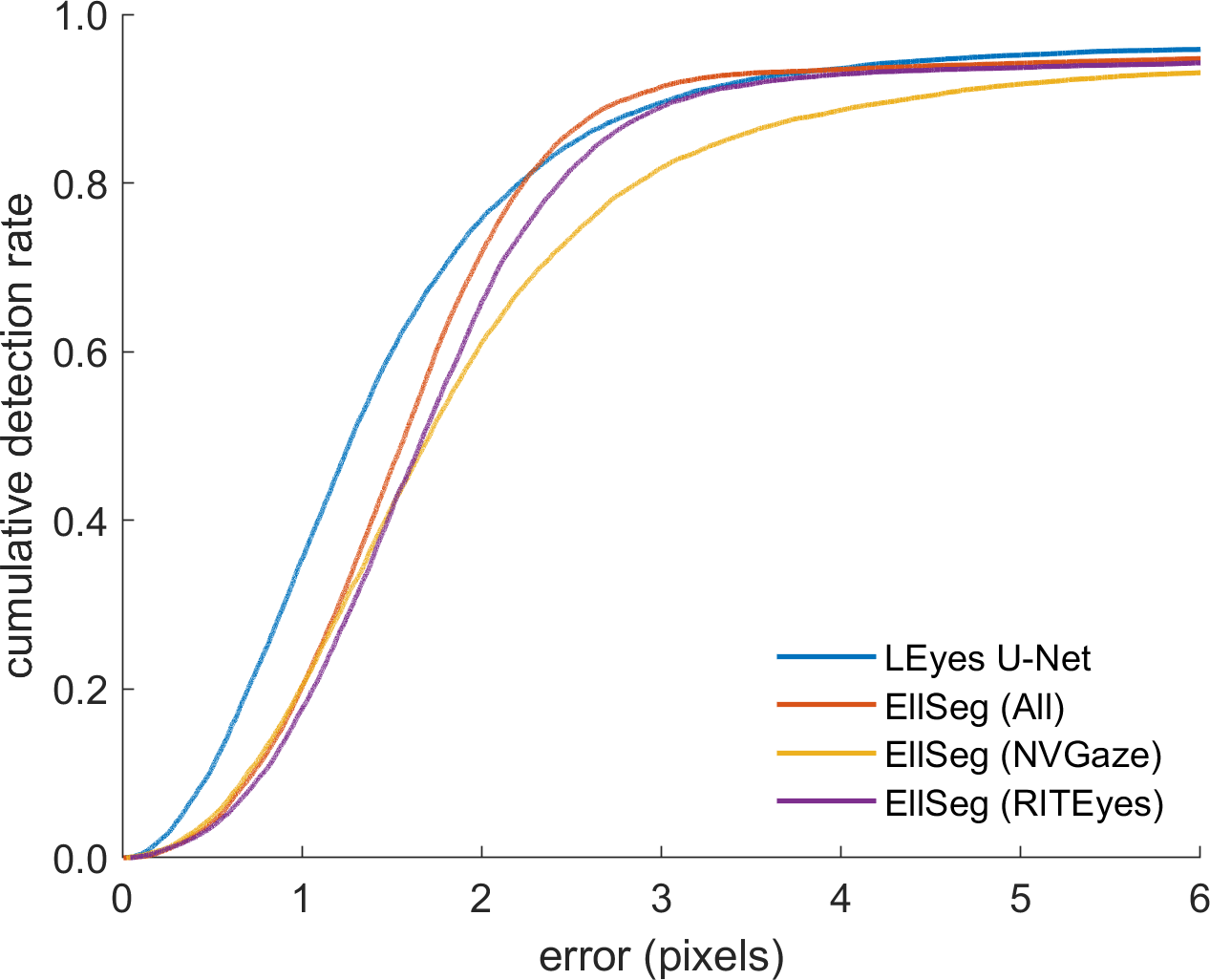}}
\\
\subfloat[]{\includegraphics[height=0.31\textwidth]{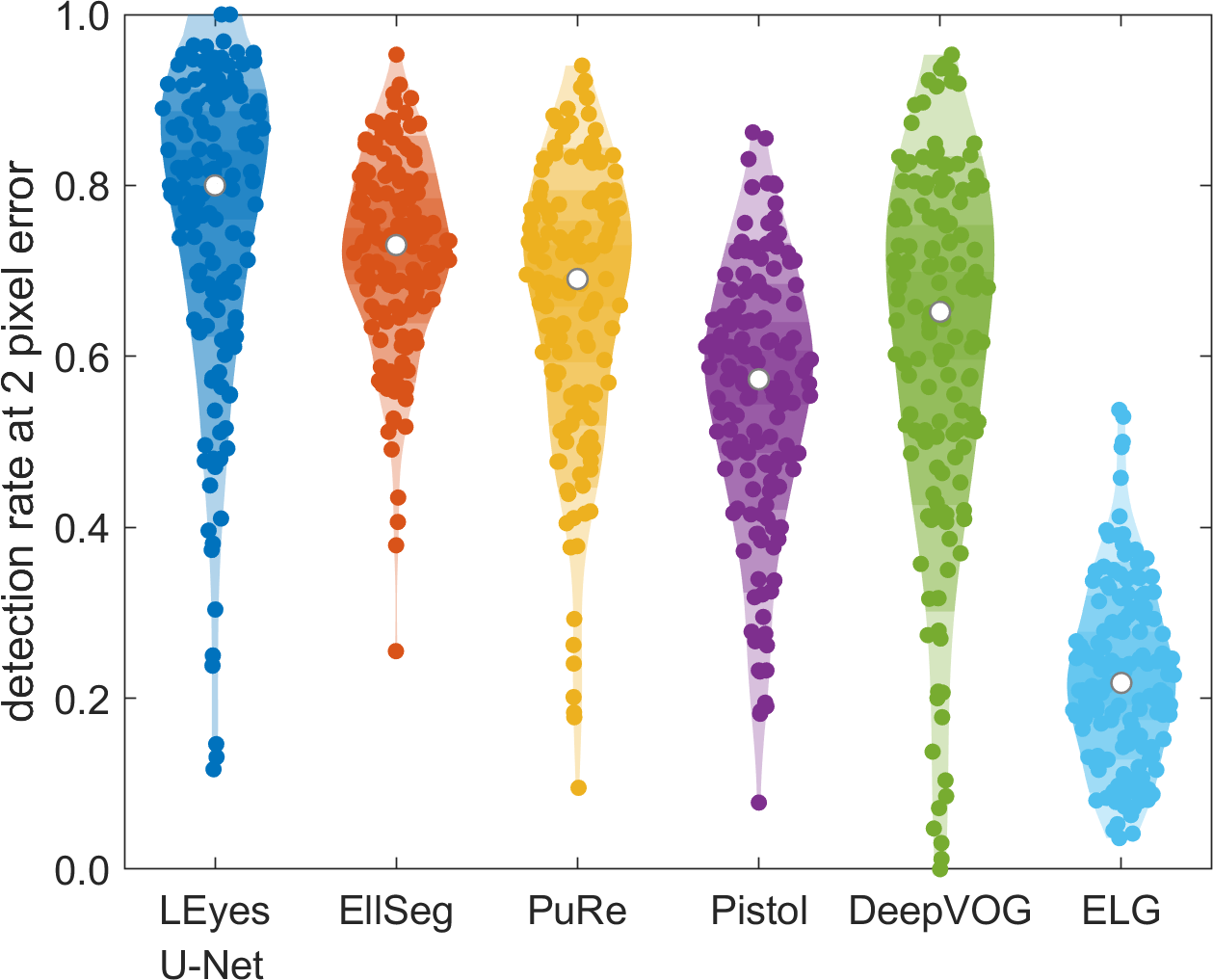}}
\hspace{1em}
\subfloat[]{\includegraphics[height=0.31\textwidth]{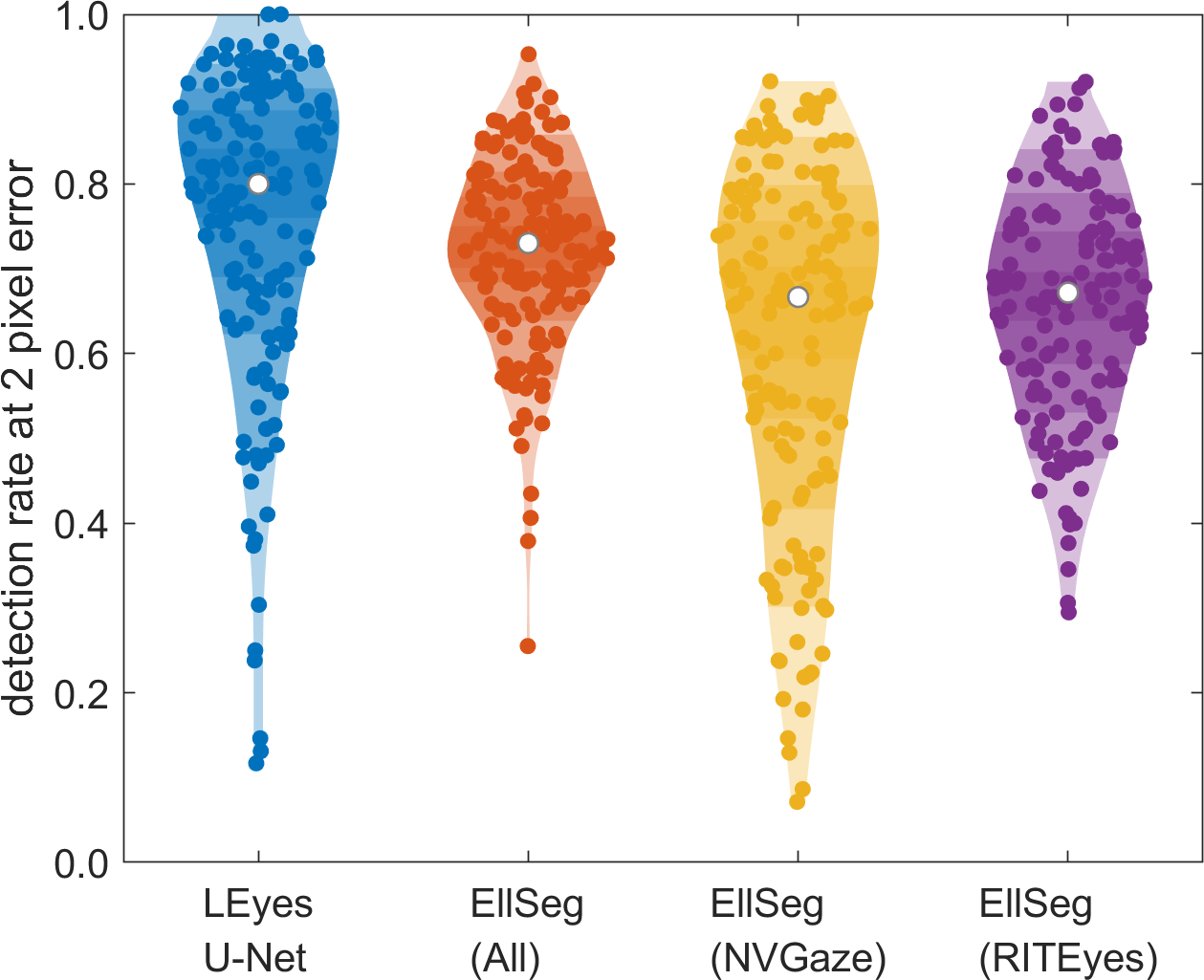}}
\caption{\textbf{A.} We compare the cumulative detection rate on the OpenEDS 2019 dataset of a U-Net model trained using the LEyes method at different pixel errors against PuRe~\cite{santini2018pure}, Pistol~\cite{fuhl2023pistol}, DeepVOG~\cite{yiu2019deepvog}, ELG~\cite{park2018learning}. \textbf{B.} We make special comparisons with several models trained using the EllSeg Framework ~\cite{kothari2020gaze,Ellseg_gen}. \textbf{C \& D:} The corresponding violin plots for panels A and B respectively, showing the detection rate at 2 pixel error for each participant in the testing set achieved by LEyes compared with the aforementioned models.} 
\label{fig:EDS2019}
\end{figure*}

Estimation accuracy in eye-tracking applications is often evaluated using the cumulative detection rate, which shows how much of the pupil locations estimated by a method are within a given distance from the ground truth pupil center~\cite{kim2019nvgaze, kothari2022ellseg, fuhl2017pupilnet, santini2018pure}. Performance is often specifically assessed as the percentage of images for which the pupil location was estimated within a 5-pixel distance from the ground truth ~\cite{kim2019nvgaze, fuhl2017pupilnet}. However, recent algorithms have demonstrated superior performance on VR datasets, often reaching ceiling performance well below this 5-pixel threshold. Consequently, we have narrowed our analysis to examine performance for errors up to just 2 pixels. As illustrated in Figure~\ref{fig:EDS2019}, we achieved a 2-pixel error rate of 75.8\%, which surpasses EllSeg (model trained on all datasets) at 71.8\% and is markedly superior to Pure (65.6\%), DeepVOG (60.9\%), and Pistol (55.4\%). The violin plots in the bottom section of Figure~\ref{fig:EDS2019} indicate that the distribution of performance across participants in the testing set at the 2-pixel level for a model trained on LEyes is comparable to other models. Notably, its median value at this error level is 80\%, outperforming the next best model by 7\%.

We underscore comparisons with the different variants of the EllSeg framework~\cite{kothari2022ellseg, Ellseg_gen}, one of the few public frameworks leveraging synthetic training data, like ours. The architecture used in the the EllSeg framework, named DenseElNet~\cite{Ellseg_gen}, has a comparable number of trainable parameters (2.24 million) compared to our model. The orange line in Figure~\ref{fig:EDS2019}(B) shows the EllSeg variant trained across multiple eye datasets. Notably, OpenEDS 2019 is one of the datasets included in its training set. Astonishingly, our LEyes model still surpasses this variant, even though there is evident data leakage with 88.6\% of the training samples present in the test set. Two further comparisons of EllSeg models trained on purely synthetic datasets (RITeyes~\cite{nair2020rit} and NVGaze~\cite{kim2019nvgaze}) show that the LEyes framework consistently outperforms other publicly available models that use only synthetic data.

Taken together, the results highlight that LEyes achieves higher performance against other methods tested on the EDS 2019 dataset. In line with earlier observations on domain discrepancies and generalization in gaze estimation~\cite{Ellseg_gen,kothari2022ellseg, nair2020rit}, our results demonstrate that models exhibit optimal performance when trained on datasets analogous to their respective test distributions, something that is easily achieved within the LEyes framework. Notably, this efficacy persists even when there are discernible differences, from a human perspective, between the training and test datasets.

\subsection*{Simultaneous Pupil and Corneal Reflection Localization}
 
Pupil and CR localization can be treated as two separate problems, yet since their positions in the eye images co-vary in a systematic way, it is advantageous to consider them together. In this section, we present a new P-CR eye tracking pipeline, trained entirely using LEyes images. Eye trackers often use several light sources to guarantee at least a pair of reflections for every gaze position, and for robust eye tracking it is necessary to reliably associate at least two corneal reflections with their specific light source across all anticipated eye movements~\cite{chugh2021detection}. Therefore, our pipeline is not only able to localize the pupil and CR centers in an eye image, but also match the CRs to specific light sources.
While previous work has developed models that locate the pupil and CRs simultaneously and perform CR matching~\cite{niu2021real,maquiling2023virnet}, we introduce a novel method that importantly streamlines the process of robust P-CR eye tracking by using the maximum value of the model output to select only the `best' two CRs. This ability of our method to robustly select CRs for gaze estimation is especially important due to the complex reality often encountered in eye images where CRs may be missing or additional, unwanted, reflections are often present. 

\begin{figure*}
    \centering
    \includegraphics[width=.9\textwidth]{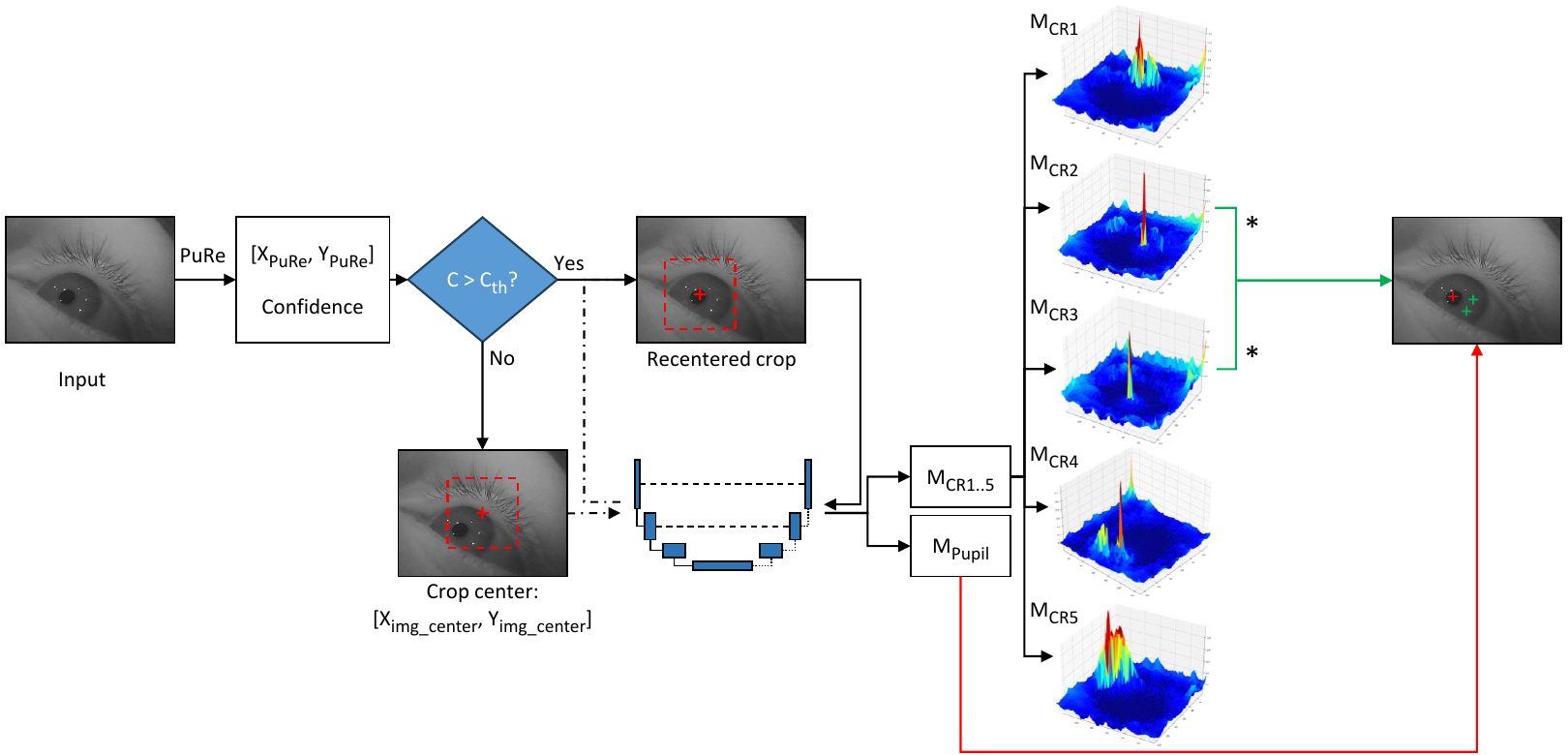}
    \caption{Flowchart of the simultaneous P-CR pipeline: Using an adaptive cropping strategy the center of the crop is determined using PuRe's pupil center prediction ($[X_{PuRe}, Y_{PuRe}]$) if the confidence metric for PuRe's prediction ($C$) is above a given confidence threshold ($C_{th}$), otherwise, the crop is determined by the pupil prediction of the LEyes-trained model given a naive center crop ($[X_{img\_center}, Y_{img\_center}]$). The pupil-centered crop is passed through the model, which outputs logits representing likely feature locations for each prediction, illustrated here as heat maps ($M$) for both the pupil ($M_{Pupil}$) and for each CR ($M_{CR1...5}$ in this example). For each CR map, the highest value is located. These peaks are compared between maps and the two highest values across all the maps determine which CRs are selected. The asterisks signify which maps contain the two highest values in this example. However, if the exclusion criteria are met, the image is deemed invalid (see text).}
    \label{fig:pureleyes_pipeline}
\end{figure*}

Through our novel pipeline, illustrated in Figure~\ref{fig:pureleyes_pipeline}, we aim to demonstrate the power of LEyes in such challenging scenarios. First, since LEyes requires input images of a certain size that contain the pupil, we have developed a novel adaptive cropping strategy. Second, we demonstrate the success of our strategy to select the two `best' CRs in an eye image.

We test our new P-CR pipeline using two VR datasets. First, we use a dataset compiled by Chugh, et al. (2021)~\cite{chugh2021detection} which contains eye images of 15 participants captured from a VR headset with an eye tracking attachment. The dataset includes manually annotated $(x, y)$ coordinates for the pupil-center and centers of the CRs. Second, we test on the OpenEDS 2020 Challenge dataset (OpenEDS 2020)~\cite{palmero2021openeds2020}, which consists of 200 participants and includes manually annotated segmentation labels for the pupil for 5\% of the data, amounting to a total of 2605 images. The images were captured at a frame rate of 100 Hz under controlled illumination using a VR headset. Since only pupil, but no CR, annotations are provided with the Open EDS 2020 dataset~\cite{palmero2021openeds2020}, we provide illustrative examples instead of a quantitative comparison. Through these examples, we want to highlight that our model appears to provide accurate predictions also for CRs in this dataset, despite have more CRs (eight instead of five) with a different spatial configuration compared to the Chugh, et al. (2021) dataset. Figure~\ref{fig:probmaps} (bottom) illustrates how our model performs on a representative selection of eye images from the OpenEDS 2020 dataset. Predictions from all eye images are available in the repository associated with this paper.

\subsubsection*{Adaptive Cropping Strategy}
Before inputting the eye image into the model, it needs to be cropped to the input size expected by the model in such a way that the pupil is in the crop. A naive cropping strategy assumes that the pupil is in the center of the eye image. However, this is not always the case and such a crop may exclude parts of, or even the entire pupil from the crop. To solve this challenge, we employ PuRe~\cite{santini2018pure}, a well known lightweight open-source pupil detection method based on ellipse fitting, to create a 128$\times$128 pixel image centered on its detected pupil center. We ran PuRe over the two datasets and found that PuRe had average pixel errors of 13.23 and 20.77 in the OpenEDS 2020 and Chugh et al.~2021 datasets, respectively. Next, providing these crops based on PuRe's pupil center estimate to LEyes still yielded high average errors of 11.0 and 7.76 pixels in those datasets due to cases where PuRe failed to locate the pupil. Therefore, we adopted an adaptive cropping strategy using PuRe's confidence metric. This confidence, ranging between 0 and 1, is based on various metrics outlined in detail in the paper~\cite{santini2018pure}, with 0 indicating a poor ellipse outline. In our cropping method, if PuRe's confidence is larger than or equal to a threshold, the crop used as input to the LEyes U-Net is based on PuRe's pupil center estimate. If the confidence is below this threshold, we instead use the naive center crop on the image with no guarantee that the pupil will be present. This hybrid cropping strategy significantly improves our model accuracy. For the OpenEDS 2020 dataset the lowest average pupil error was 4.18 pixels, achieved when using a confidence threshold of 0.90 (the largest average pixel error was 4.76 for confidence thresholds between 0.50--0.95). For the Chugh et al.~2021 dataset, an average pixel error of 4.15 pixels was achieved at the 0.70 confidence threshold (largest error 5.11). 



\subsubsection*{Selecting the 'best' CRs using model output}

\begin{figure*}
    \centering
    \includegraphics[width=.9\textwidth]{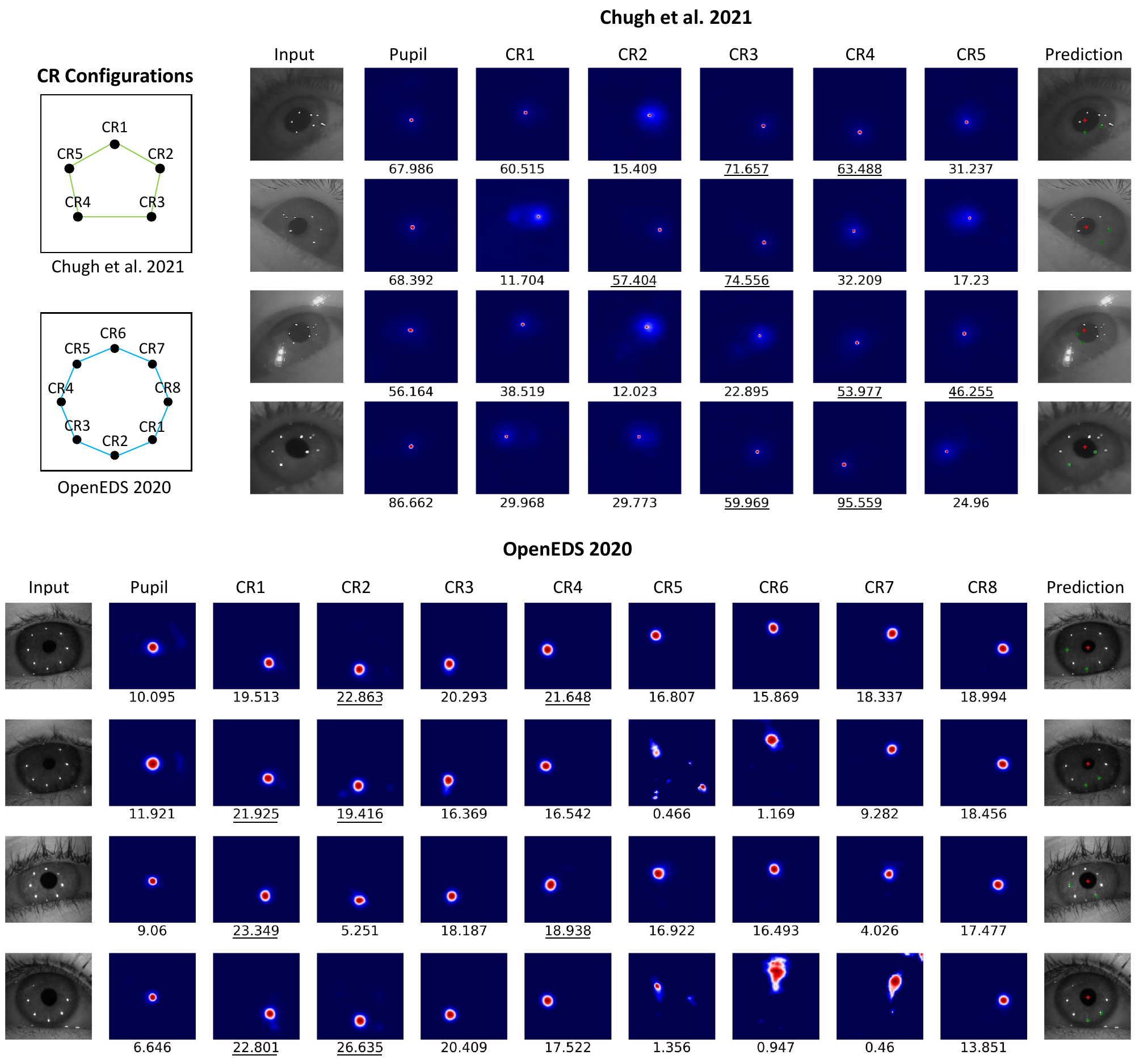}
    \caption{Heat maps for both the Chugh et al.~2021 dataset and the Openeds 2020 dataset. The maximum of the corresponding logit value is shown under each heat map. In the Chugh et al.~2021 dataset, the labeling of the CRs starts at the top-most IR reflection and then proceeds clockwise (top right). In the OpenEDS 2020 dataset, the labels used when training the model start at the lower right CR and proceed clockwise. Our algorithm selects the two highest logit values from the CR maps along with the pupil value for a complete robust P-CR pipeline. The last column shows the prediction locations of the centers of the pupil and selected CRs on the corresponding eye image.}
    \label{fig:probmaps}
\end{figure*}

The LEyes U-Net model takes a grayscale eye image as input and produces output maps for each feature (the pupil and each CR) that correspond to the confidence the model has that a given feature's center is located at a given position in the input image. We will represent these unnormalized output values, which we will refer to as logit values, in the form of a heatmaps. The maximum value of each heatmap corresponds to where the model is most confident of the prediction for the pixel location of each eye feature's center. To robustly select the two CRs the model is most confident about, we choose the two CRs with the highest corresponding logit values across the output heatmaps. Figure~\ref{fig:probmaps} shows the heat maps of each CR and their associated max values derived from real eye images from both datasets along with the predicted locations of the selected CRs overlaid onto the eye image. To exclude eye images clearly unsuitable for eye tracking, for instance images that contain a blink or when both cropping strategies failed to capture the pupil, our method excludes images that fail to produce at least two heatmaps where the max values are greater than or equal to one. 

To assess our model, we compared its average pixel error to the CR annotations in the Chugh et al.~2021 dataset~\cite{chugh2021detection}. They achieved successful matches of at least two CRs within five pixels for 91\% of the images in their test set and an average error of 1.5 pixels on these images.
It is worth mentioning that Chugh et al.~2021 had to sacrifice 88\% of the dataset for both training and validation of the model~\cite{maquiling2023virnet}, so their results include only a small part (12\%) of the whole dataset.
In contrast, since LEyes is trained on synthetic images, we can evaluate our model on the entire dataset. Therefore, direct comparisons between the two models are not straightforward since they are evaluated on a different number of images and use different exclusion criteria. To make the results more comparable, we apply our exclusion criterion that the maximum value of at least two heat maps is larger than one in conjunction with Chugh et al. (2021)'s criterion that evaluates model performance only on the images where the predicted locations of at least 2 CRs were less than 5 pixels away from the ground truth. Using these criteria, our model exhibited an average pixel error of 1.59 across all the CRs. Focusing solely on the best two CRs, this error was reduced by 18\% to 1.30 pixels. Further, using both exclusion criteria we retain 70\% of images from the dataset.

\subsection*{High-Resolution Gaze Tracking}

\begin{figure*}
    \centering
    \includegraphics[width=.9\textwidth]{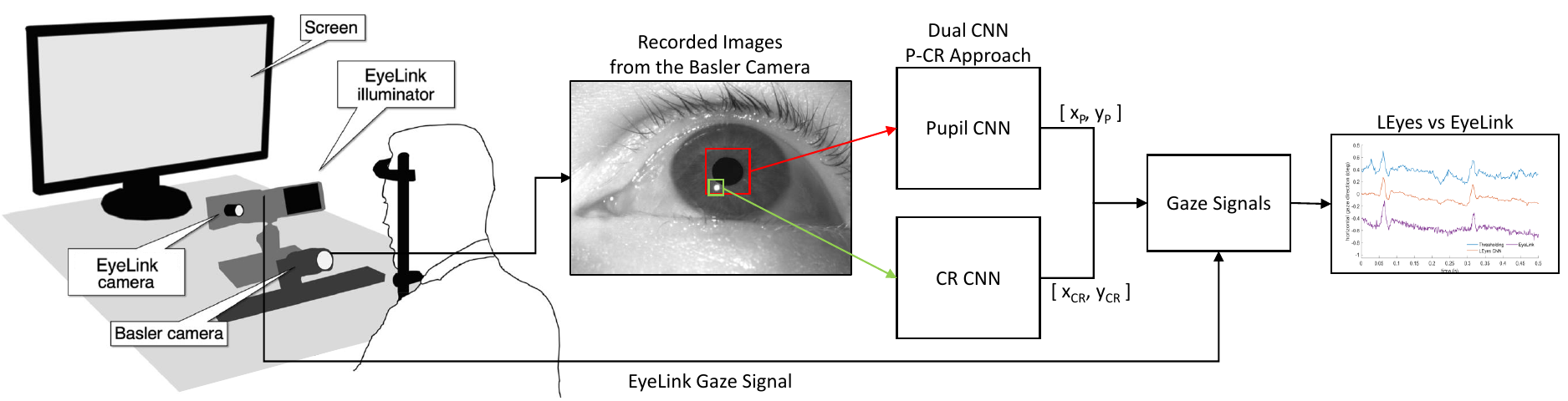}
    \caption{Experimental setup: In a co-recorded setup we acquire eye images from the FLEX setup and gaze signals from the Eyelink 1000 Plus. We analyzed the eye images which we recorded from expert participants using a dual CNN approach. The pupil CNN localized the pupil center, while the CR CNN localized the center of the CR located in the eye image. Both CNNs achieved sub-pixel pixel error. Image of co-recording setup adapted from~\cite{nystrom2023amplitude}.}
    \label{fig:flex}
\end{figure*}

The Pupil-Corneal Reflection (P-CR) eye tracking method, often employed in controlled lab settings for gaze estimation, requires accurate identification of both the pupil and Corneal Reflections (CRs)~\cite{hooge2016pupil, nystrom2023amplitude, fuhl2017pupilnet}. When estimating the smallest and fastest of eye movements, an eye tracker with high spatial and temporal resolution is required. This typically requires sub-pixel localization of the pupil and CR(s).

To address these requirements, we developed a dual Convolutional Neural Network (CNN) model trained on LEyes images. One CNN focuses on locating the pupil center, while the other locates the CR center; an illustration of our setup is in Figure~\ref{fig:flex}.
Our model was compared with traditional thresholding methods, the LEyes U-Net model used in OpenEDS 2019 but with different parameters used in generator to account for the dataset, and a state-of-the-art commercial eye tracker (SR Research EyeLink 1000 Plus).

The data for this high-resolution study was captured in a co-recorded experiment using our custom-built FLEX system \cite{nystrom2023amplitude,hooge2021pupil} and the EyeLink 1000 Plus eye tracker. Such co-recording was required since the eye images captured by the EyeLink are not accessible and a direct comparison of its image processing to the LEyes method is thus not possible. The EyeLink's illuminator was used to deliver illumination for both systems. This setup resulted in eye images containing a single CR. Both the FLEX system and the EyeLink acquired data at 1000 Hz. Since the focus of this comparison is on eye tracking signal quality, data was recorded from 4 expert participants who performed a series of fixation and saccade tasks during eight minutes. To provide additional variation in the luminance profiles of the eye images, and thereby test the robustness of our model, the four participants were recorded a second time with the FLEX system configured to a sampling rate of 500 Hz. The captured eye images were brighter at this lower sampling rate due to the longer possible exposure time.

\begin{figure*}
    \centering
    \includegraphics[width=.60\textwidth]{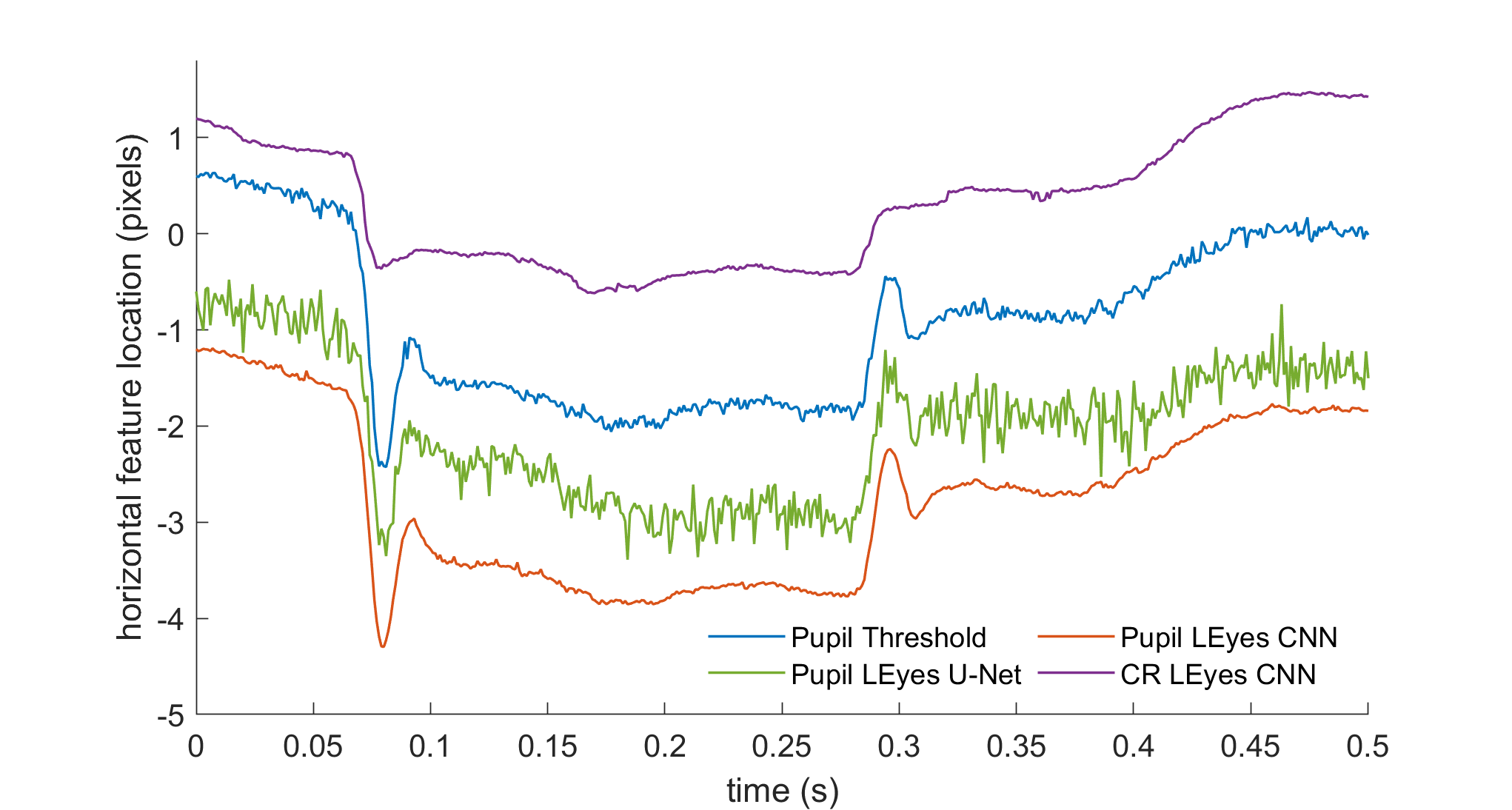}
    \caption{Representative segment of pupil and CR center locations derived from 1000 Hz eye images. The pupil center was determined using three different methods; thresholding (blue), a U-Net trained using the LEyes framework and derived from the EDS2019 U-Net (green), and a CNN trained for pupil center localization using LEyes images (red).}
    \label{fig:high_res_bridge}
\end{figure*}

The eye images captured by the FLEX system were first processed using a standard thresholding operation~\cite{nystrom2023amplitude} to provide an initial localization of the pupil and CR centers. We then took 180x180 pixel crops centered on the pupil and the CR features from the original images and fed these to the LEyes CNNs. As shown in Figure~\ref{fig:high_res_bridge}, both thresholding and in particular the LEyes CNNs provided a significant improvement in stability of the pupil signal compared to using the LEyes U-Net model, which therefore was omitted from further analysis.

\begin{figure*}
    \centering
    \includegraphics[width=.8\textwidth]{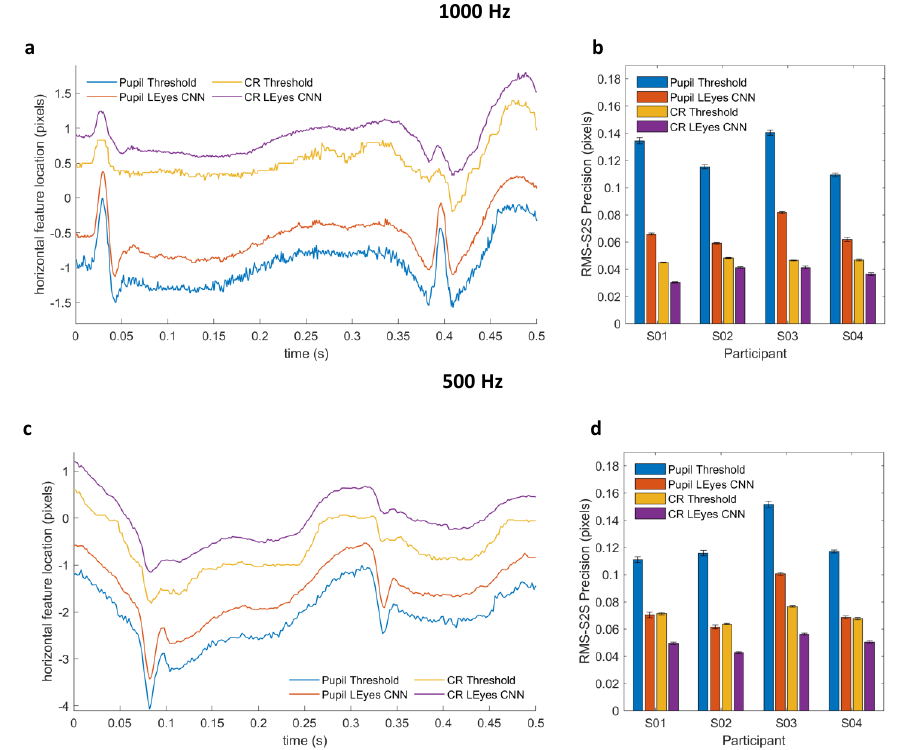}
    \caption{CR and pupil center signals. Left column: representative segment raw pupil and CR center signals derived from eye images recorded at 1000 Hz (a) and 500 Hz (b). Right column (panels b and d): an RMS precision comparison between the thresholding and LEyes CNN methods for the pupil and CR signals on all data of four participants. Error bars depict standard error of the mean.}
    \label{fig:rawsignal}
\end{figure*}

Example raw pupil and CR signals resulting from the thresholding operations and the LEyes CNNs are shown in Figures~\ref{fig:rawsignal}a (1000 Hz) and \ref{fig:rawsignal}c (500 Hz). As can be readily appreciated, the sample-to-sample variation in both the pupil center and the CR center signal is lower for the LEyes method than for the standard thresholding method for data acquired at both sampling rates. To formalize this observation, the precision in the form of root mean square of sample-to-sample deviations in the signal (RMS-S2S \cite{HoNyMu2012,niehorster2020impact,niehorster2020characterizing}) was computed across the dataset and plotted in Figures~\ref{fig:rawsignal}b (1000 Hz) and \ref{fig:rawsignal}d (500 Hz). This analysis confirms that for both the 1000 and the 500 Hz data sets, the LEyes CNNs consistently demonstrated superior precision (lower values) than the thresholding method. 

\begin{figure*}
    \centering
    \includegraphics[width=1\textwidth]{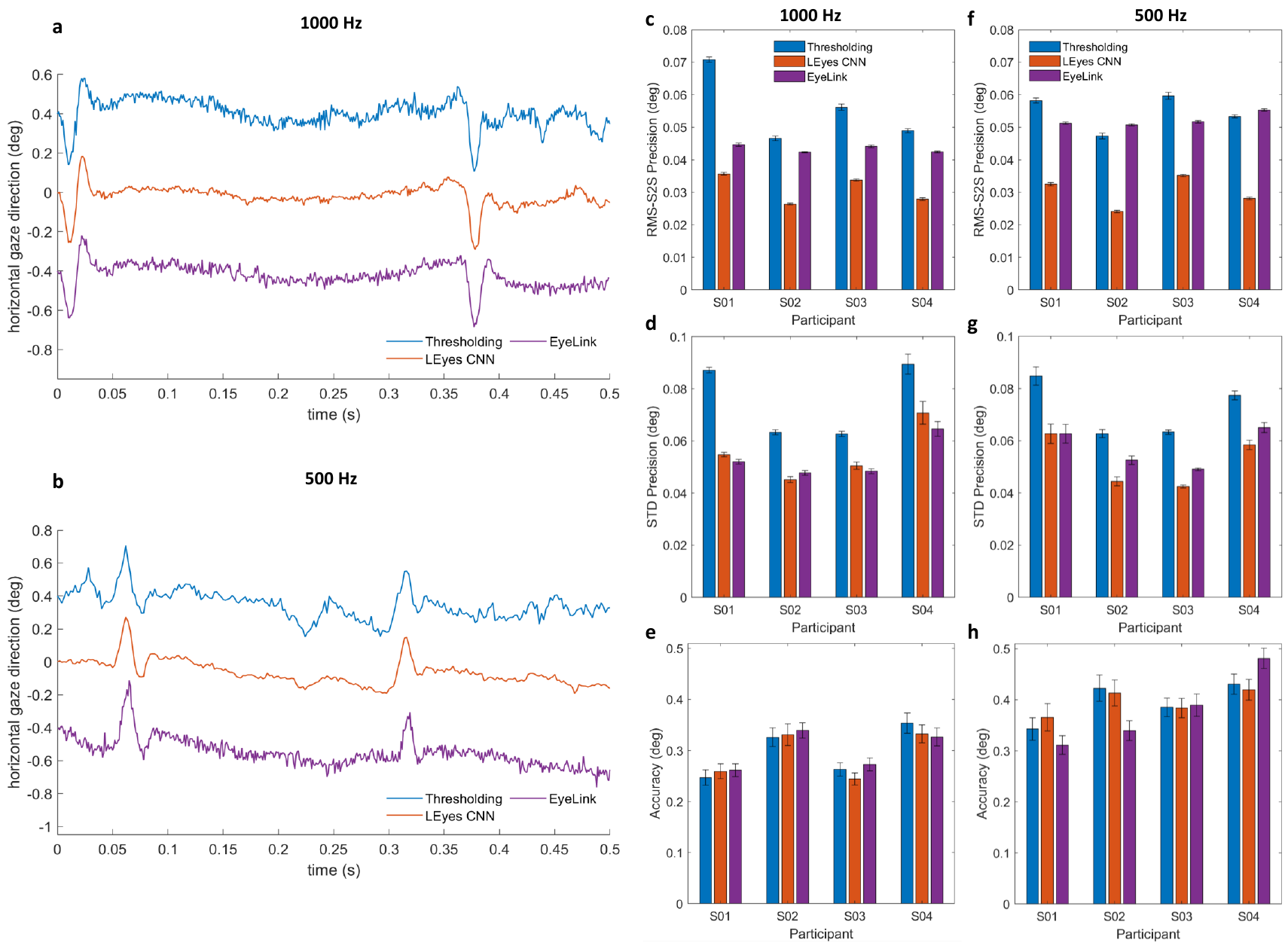}
    \caption{Calibrated gaze signals. Left column: representative segment of calibrated P-CR signals derived from 1000 Hz data (a) and 500 Hz data (b) as derived from pupil and CR center locations determined using either thresholding or the dual LEyes CNN strategy, along with the EyeLink. The signals in both panels contain two small saccades and have been vertically offset for clarity. Further, an RMS precision, STD precision and an accuracy comparison for the 1000 Hz data (middle column, panels c--e) and the 500 Hz data (right column, panels f--h) between the three gaze tracking methods on data of all participants are shown. Error bars depict standard error of the mean.}
    \label{fig:gazesignal}
\end{figure*}

Researchers using eye tracking are rarely interested in the individual pupil and CR signals, but instead use the gaze signal derived from them. Does the improved precision of the pupil and CR center signals lead to an improved gaze signal? To examine this, we derived P-CR gaze signals using pupil and CR centers estimated by thresholding or by the LEyes CNNs and compared both with the gaze signal delivered by the EyeLink. Each signal was calibrated using standard methods and example segments are plotted in Figures~\ref{fig:gazesignal}a (1000 Hz) and \ref{fig:gazesignal}b (500 Hz). Again, it can be readily appreciated that the gaze signal derived from the LEyes CNNs is smoother and more stable than that derived from standard thresholding operations or delivered by the EyeLink. To quantify this observation, we computed the RMS-S2S precision of these signals which quantifies short-timescale smoothness, as well as the STD precision \cite{niehorster2020characterizing} which quantifies the spatial spread of the signal and indicates its stability. These evaluations are presented in Figures~\ref{fig:gazesignal}c--e (1000 Hz) and \ref{fig:gazesignal}f--h (500 Hz). This analysis confirms that the dual LEyes CNN method consistently demonstrated superior RMS-S2S precision (lower values) than the thresholding method and the results from the Eyelink 1000 Plus. It is important to note that all methods processed each video frame independently, without using any temporal information from preceding or future frames. Thus, the increased precision seen in the CNN method cannot be attributed to the use of temporal information~\cite{niehorster2021apparent,niehorster2020characterizing}. The signal stability (STD precision) achieved by the LEyes method was on par with the EyeLink for the 1000 Hz dataset and slightly better than the EyeLink for the 500 Hz dataset, and consistently outperformed the thresholding method for both datasets. The accuracy achieved did not systematically differ between the three methods, indicating that the gains in precision did not come at the cost of reduced accuracy.

\section*{Discussion}

We developed a novel framework named LEyes for training gaze estimation algorithms, achieving cutting-edge results for both virtual reality (VR) and high-resolution, lab-based eye-tracker setups. LEyes outperformed other methods in a pupil center localization task by a margin of at least 4\%. In a high-resolution setting, LEyes exceeded the performance of the industry-standard EyeLink 1000 Plus eye tracker across two lighting conditions in a co-recorded experiment. Additionally, we introduced a novel LEyes-trained P-CR pipeline that both simplifies and improves CR detection by considering only the two best CRs in the recorded image. Overall, our results emphasize both the accuracy and flexibility in design of the LEyes framework, highlighting its applicability across gaze estimation applications.

LEyes has the potential to be a game-changer for the many companies and startups attempting to enter the VR and eye-tracking space. LEyes enables these companies to bring their devices to market without the necessity of collecting or purchasing potentially millions of eye images from a third party, alleviating both the costs and hurdles related to data acquisition. This opens up a streamlined path to market, making it an attractive option for emerging companies. In an academic setting, LEyes significantly reduces the amount of data required to conduct an eye-tracking study that uses a deep learning model to analyze the data, by eliminating the need to sacrifice recorded data for model training and validation, resulting in both time and cost savings. For example, our model was able to run inference on the entirety of the Chugh et al.~2021 dataset, while the original paper used 88\% of the data for both training and validation and were thus left with only 12\% for evaluating there model~\cite{chugh2021detection, maquiling2023virnet}. Furthermore, using Python, LEyes offers an alternative to the challenging task of creating photorealistic synthetic data. Many researchers may not possess the skills, time, or resources to access and use software platforms like Blender or Unity3D. Finally, when combined with the FLEX system which has a hardware cost of about \$1000 USD, LEyes offers a low cost and open source alternative to the EyeLink 1000 Plus. 

Our study has limitations: First, the models were trained on simulated data but tested on real data. We did not investigate any potential learning differences between synthetic and real eye datasets. Future research may benefit from analyzing these differences to further improve the quality of the synthetic data generation. Second, we aim to explore "Domain Adaptation" techniques such as fine-tuning LEyes-trained models with real eye images to assess performance impact. Third, the low participant count in our high-resolution experiment, while sufficient for our purpose of demonstrating the power of a LEyes trained model, potentially limits the generalization of our findings for this specific test. Despite seeing promising results with the LEyes framework and good generalizability across large participant samples in the other tests, recruiting a broader participant base that encompasses both experts and novices can be seen as a worthwhile further study.

Prior to LEyes, the development of gaze estimation algorithms using machine learning was confined to those who possessed the resources to amass large annotated datasets or the technical expertise and large computational resources to generate synthetic data. With LEyes, the training of deep learning models for gaze estimation has become easily accessible to everyone, democratizing the field and opening new avenues for exploration and application.

\section*{Methods}

We have made the code for the various simulations and model training regimes described below, as well as the trained models and the code for evaluating the model on the various real image data sets available at the following link: \url{https://github.com/dcnieho/Byrneetal_LEyes}.

\subsection*{Generating Light Simulations}

Five different simulations modeling the light distribution of eye images were used for training the U-Net for OpenEDS 2019, the U-Net models with attention used on the OpenEDS 2020 and Chugh et al.'s 2021 datasets and the CR and pupil CNNs. Here we first present features shared between these simulations, and then detail the individual simulations in order of complexity. The full code to generate LEyes images is available at our GitHub Repository linked to this paper. 

\subsubsection*{Common features}

Following previous work \cite{byrne2023CRCNN,maquiling2023virnet}, we developed simulated images that model the light distributions of the relevant aspects of an eye image that the given model would have to deal with during inference. Blob-like features, such as the pupil and CR were modeled as 2D Gaussian distributions using the equation
\begin{equation}
G(x,y) = Ae^{-a(x - x_c)^2 - b(x - x_c)(y - y_c) - c(y - y_c)^2 },
\label{eq:Gauss}
\end{equation}
where
\begin{align}
    a &= \frac{\cos(\theta)^2}{2\sigma_\alpha^2} + \frac{\sin(\theta)^2}{2\sigma_\beta^2},\\
    b &= \frac{\sin(2\theta)}{4\sigma_\alpha^2} - \frac{\sin(2\theta)}{4\sigma_\beta^2},\\
    c &= \frac{\sin(\theta)^2}{2\sigma_\alpha^2} + \frac{\cos(\theta)^2}{2\sigma_\beta^2},
\end{align}
and where $\theta$ is the orientation of the 2D Gaussian and $\sigma_\alpha$ and $\sigma_\beta$ its spread along the minor and major axes, respectively.

The luminance of the pupil was determined per simulation by analyzing the eye images on which inference would be run, while the luminance of a CR was always set to full white. Regardless of the Gaussian amplitude $A$ of the feature, which was varied to create differently steep edges, the minor and major axis radii of the luminance plateau in each feature (the dark part of a pupil, or the bright part of a CR) were kept constant by parameterizing 
\begin{equation}
    \sigma_r = r/\sqrt{-2\log \frac{1}{A}}, r \in \{\alpha,\beta\}.
\end{equation}

To create the final simulated image, first the relevant features were layered onto a background luminance distribution that differed between simulations. These layers were then collapsed into a single image by subtracting dark features (such as pupils) from the background, and by adding bright features to the collapsed image of the preceding layers using the operation $max(image, background)$. Pixel noise was added to the final image by adding a value from a Gaussian distribution $X \sim \mathcal{N}(0,\,\sigma_n^{2})$ to the image that was drawn independently for each pixel. Finally, the resulting image was limited to the range $[0, 255]$, scaled to the range $[0,1]$ and discretized to 256 levels, corresponding to 8-bit camera images.

\subsubsection*{CR 500 Hz \& CR 1000 Hz}

The CNN for CR center localization used for the 500 Hz data was the same as presented in previous work\cite{byrne2023CRCNN}. As such, only the key points of this simulation are described.
Circular CRs ($\sigma_\alpha=\sigma_\beta \in [1,30]$, $A\in [2,20000]$) were placed on a background that was made up of two parts, divided by a randomly oriented straight line representing the pupil-iris border that passed close to the CR. On one side of the line the background was dark, with a luminance drawn from an exponential distribution with its scale parameter set to 10 pixel intensity values, and offset 1. The other part of the background was middle grey (pixel intensity value $L_{CR}=128$). The standard deviation of image noise was varied per generated image, with $\sigma_n \in [0,30]$.

The simulations used for training the CNN for determining CR centers in the 1000 Hz eye videos were identical to those used for the 500 Hz data, except that the middle-grey part of the background varied in luminance between $L_{CR} \in [32,153]$.

For both the 500 Hz and the 1000Hz models, in the second stage the location of the CR center was constrained to a range spanning 1.5 pixels around the image center.

\subsubsection*{Pupil 500 Hz \& Pupil 1000 Hz}

The simulated light distributions used for training the CNN for locating pupil centers differed from the simulations for the CR CNNs in a few ways. First, the simulated images contained a 2D Gaussian representing the darker pupil. Second, the images contained one or multiple bright 2D Gaussians representing CRs that were randomly positioned and could thus overlap the pupil. Third, instead of a background consisting of dark and grey segments separated by a straight line, the background now consisted of a uniform field at a range of grey levels, representing the iris at various illumination levels.

Specifically, a randomly oriented dark 2D Gaussian with minor axis radius $\alpha_p \in [20,60]$ pixels, major axis radius $\beta_p \in [1\alpha_p,1.3\alpha_p]$ and amplitude $A_p \in [2,20000]$ was used to represent the pupil. Its luminance $L_p$ was drawn from an exponential distribution with a scale parameter of 10, and offset 1.
Between 1 and 4 corneal reflections (CRs) were generated with minor axis radius $\alpha_c \in [4,12]$ and major axis radius $\beta_c \in [1\alpha_c,1.1\alpha_c]$ and $A_c \in [2,20000]$ and randomly positioned. Overlap between CRs was avoided by removing CRs whose center location was closer to another CR than 1.25 times the sum of the major axis radii of the two CRs, and replacing it with a new randomly positioned CR. The background luminance level representing the iris was $L_{background} \in [64,179]$ pixel intensity values. The standard deviation of image noise was varied per generated image, with $\sigma_n \in [0,30]$.

The simulations used for training the CNN for determining pupil centers in the 1000 Hz eye videos were identical to those used for the 500 Hz data, except that the background luminance level representing the iris was $L_{background} \in [32,153]$ pixel intensity values to encompass the iris luminance values in the darker 1000 Hz eye images.

For both the 500 Hz and the 1000Hz models, in the second stage the location of the pupil center was constrained to a range spanning 1.5 pixels around the image center and only 1 randomly positioned CR was generated.

\subsubsection*{U-Net for OpenEDS 2019}

In order to ensure that the U-Net reliably detects the pupil and not the iris, the simulations used for training the U-Net contained several more features than those for the pupil CNN. Firstly, a bright background representing the sclera with luminance $L_s \leftarrow \mathcal{N}(217,\,26)$ was generated. On top of this an iris was generated as a randomly positioned and oriented ellipse ($\alpha_i \in [30, 42.5]$ and major axis radius $\beta_i \in [1\alpha_i,1.3\alpha_i]$ and $L_i \leftarrow \mathcal{N}(77,\,16)$) rendered with an edge modulated by a raised cosine function over a range of between $[8,20]$ pixels. Then an irregularly shaped collarette was generated close to the center of the iris consisting of between 13 and 24 vertices arranged around the collarette center at an average distance $r_{col} \in [.3\beta_i,.6\beta_i]$, with the individual distance of vertices varied between $[0.05r_{col},0.2r_{col}]$. The resulting polygon was upsampled to five times the number of vertices using periodic cubic spline interpolation to create a shape with a smoothly varying edge, and the resulting polygon was rendered at luminance $L_{col}=[1.25L_i,1.6L_i]$ with an edge modulated by a raised cosine function over a range of between $[1,4]$ pixels.

On top of this were layered a randomly positioned and oriented pupil (minor axis radius $\alpha_p \in [10, 30]$ and major axis radius $\beta_p \in [1\alpha_p,1.3\alpha_i]$, $A_p \in [2,2000]$ and $L_p \leftarrow \mathcal{N}(34,\,15)$) and between 1 and 8 randomly positioned and oriented CRs (minor axis radius $\alpha_c \in [0.8,4]$ and major axis radius $\beta_c \in [1\alpha_c,1.4\alpha_c]$, $A_c \in [2,20000]$ and $L_{CR}=255$), again avoiding overlap. The standard deviation of image noise was varied per generated image, with $\sigma_n \in [0,15]$.

\subsubsection*{U-Net for Chugh et al. 2021 dataset}

We use a simulation that improves on previous work~\cite{maquiling2023virnet} to perform pupil and CR localization and CR matching. The pupil is represented by a randomly oriented dark 2D Gaussian with a minor axis radius $\alpha_p \in [6, 22.5]$ pixels, major axis radius $\beta_p \in [1\alpha_p, 1.3\alpha_p]$ and amplitude $A_p \in [200, 100000]$. Its luminance $L_p$ is drawn from an exponential distribution with a scale parameter of 10 and offset 1. %
Five randomly oriented CRs are generated, each having a random minor axis $\alpha_c \in [1, 2.5]$ pixels, a random major axis $\beta_c \in [\alpha_p, 1.1\alpha_p]$, and a random amplitude $A_c \in [200, 100000]$. Each CR has a drop-out rate of 16\%. %
Between 1 and 5 spurious (non-CR) reflections may randomly appear in the image. These are generated in the same way as CRs, each with a random minor axis radius $\alpha_s \in [1, 2.5]$ pixels and random major axis radius $\beta_s \in [\alpha_s, 2.5\alpha_s]$. The location of each spurious reflection is generated using a rejection sampling method with an inverted Gaussian ($1 - G(x, y)_p$, c.f. Eq \ref{eq:Gauss}) to make them less likely to appear near the pupil center. 
A grayscale gradient background was created by drawing two random values from a luminance range of $L_{background} \in [63,178]$ and smoothly varying the luminance from one side to the other along a random axis. This is to prevent the model from interpreting any dark part of the image as part of the pupil. The standard deviation of image noise was varied per generated image, with $\sigma_n \in [0,30]$.

As this model not only performs pupil and CR center localization but also matching of CRs to specific illuminators, the positions of the CRs need to follow the same pattern as in the real dataset. Specifically, for Chugh et al.'s 2021 dataset \cite{chugh2021detection}, this involves five IR lights that project to a house-shaped polygon that is usually close to the pupil. The polygon is modeled as a rectangle with an additional vertex above the middle of its top edge. The rectangle's base width is randomly sampled from $w \in [0.1d, 0.45d]$ where $d = 128$ pixels, the length of one side of the synthetic image. The rectangle's height is sampled from $[0.5w, 0.6w]$, and the height of the roof from $[0.2w, 0.5w]$. The polygon is randomly rotated between $\pm [0, 45]$ degrees.

In order for the model to learn the matching correctly, the CR positions are always calculated in a certain order, starting from the topmost position and moving clockwise. Training this model was performed in two stages (see below). In the second stage, the maximum number of spurious reflections that could appear in the image is reduced to 3, the dropout probability for individual CRs is reduced to 10\% and the range of rotation is reduced to $\pm [0, 35]$.

\subsubsection*{U-Net for OpenEDS 2020}

We reuse the simulation created for Chugh et al.~2021 dataset \cite{chugh2021detection}, adjusting the polygon so that it has eight vertices corresponding to the eight IR lights in the dataset, starting from the bottom-right CR and moving clockwise. The polygon's radius is randomly sampled from the range $w \in [0.15d, 0.4d]$ where $d = 128$ pixels. As the OpenEDS 2020 dataset contained forward-facing eye images, the random rotation of the polygon is reduced to the range $\pm [0, 0.57]$ degrees. Each CR has a dropout probability of 20\%. The pupil luminance $L_p$ is drawn from a Weibull distribution with a scale of 25, an offset of 18 and shape parameter of 2, while no other parameters were changed. 

\subsection*{Neural Network Architectures \& Training Regimes}

\subsubsection*{U-Net model for pupil segmentation}
For the pupil segmentation task, we utilized an off-the-shelf U-Net \cite{ronneberger2015unet} from the PyTorch Segmentation Modules library~\cite{Iakubovskii:2019}. The encoder part uses a ResNet-34 backbone \cite{koonce2021resnet} pre-trained on ImageNet~\cite{5206848}. The decoder part consists of five convolutional layers of dimensions (256, 128, 64, 32, 16). The trained U-Net model accepts grayscale images of arbitrary dimensions and produces a probability map that represents the pupil segmentation. In total, the U-Net model contains 24,430,097 trainable parameters. 

The masks output by the U-Net (range $[0,1]$) were binarized using a threshold of 0.99, and then postprocessed with OpenCV (version 4.7.0.68) in Python 3.10. Specifically, morphological operations were performed on the resulting binary masks to fill holes, and the pupil was selected based on shape and size criteria\cite{nystrom2023amplitude}. The center of mass of the blob was then computed and an ellipse was then fit to the selected blob. If the center of mass was closer than the radius of the ellipse's major axis to the edge of the eye image cutout, the cutout was recentered on the center of mass and inference run anew on this cutout.

\subsubsection*{U-Net with attention mechanism}
In both cases, we used a modified U-Net model based on previous work~\cite{niu2021real}. The encoder and decoder consist of residual modules producing a feature map with a consistent depth, only decreasing/increasing in size using down-and upsampling respectively. The U-Net contains six residual modules with a consistent channel size of 256. The output of the U-Net are passed through two convolution blocks which produce the heat maps for the CRs and the pupil respectively. The peak in the heat maps is taken as the pixel location of each eye feature center and is found with an argmax operation.

\subsubsection*{CNN models for pupil and CR localization} 

In this task, the model was trained to localize the subpixel center of an eye feature (pupil or CR). Overall, four CNNs were trained: two for localizing the pupil and CR centers in data captured at 500 Hz with the FLEX setup and another two for data captured at 1000 Hz. Each model is composed of seven convolutional layers followed by two dense layers. The CNNs for CR center localization in both 500 Hz and 1000 Hz data as well as the CNN trained to detect the pupil center in 500 Hz data have the following convolution layer dimensions: (64, 64, 128, 128, 256, 256, 512) while the pupil CNN for 1000 Hz data has wider dimensions: (128, 128, 256, 256, 512, 512, 768). The CR CNNs have dense layers with sizes of (64, 32) while the pupil CNNs both have sizes of (64, 64). The CR CNNs both have a total of 6,268,386 trainable parameters while the pupil CNN for 500 Hz and 1000 Hz data have 6,270,530 and 19,671,426 trainable parameters, respectively. Each CNN model was built within the DeepTrack 2.1 library~\cite{midtvedt2021quantitative}.

\subsubsection*{Model training regimes}
To train the U-Net for the OpenEDS 2019 dataset, we chose the AdamW \cite{loshchilov2019decoupled} optimizer with an initial learning rate set to $1e^{-4}$ and an exponential decay scheduler. The loss used is a combination of Binary Cross Entropy loss \cite{ruby2020binary}, Dice loss \cite{li2019dice}, and Focal loss \cite{lin2017focal}. During the training phase, the model was shown 1000 new simulated images per epoch and the validation set consisted of 400 pre-generated simulated images. The model training ran for 100 epochs reaching a natural plateau. 

Following our previous work \cite{byrne2023CRCNN, maquiling2023virnet}, the U-Net model used for Chugh's dataset \cite{chugh2021detection} is trained in two stages. The first stage consisted of a broader range of challenging examples, aimed at enhancing the model's robustness to large variations in eye data while the second stage consisted of images that more closely represent the images captured by the eye tracker. Similar to the U-Net model for EDS 2019, we used the AdamW optimizer with an initial learning rate of $1e^{-4}$ in the first stage and $1e^{-5}$ in the second stage, an exponential decay scheduler, and a combination of Binary Cross Entropy loss \cite{ruby2020binary}, Dice loss \cite{li2019dice}, and Focal loss \cite{lin2017focal} for the loss. The generator was first configured to present the model with 20000 unique images per epoch. In the second stage, the generator is reconfigured to show 1000 images. We let the model train for 30 epochs in the first stage and 20 epochs in the second stage. We incorporated early stopping with a patience of 30 for the first stage and 5 for the second stage. 

Similarly, the U-Net model for EDS 2020 \cite{palmero2021openeds2020} is trained in two stages. We let the model train for 500 epochs in both stages and incorporated early stopping with a patience of 30. In the first stage, a weight of 100 is added to the Binary Cross Entropy Loss, while the rest of the parameters for both the first and second stages remain the same as the EDS2020 U-Net model. In both stages, the generator was configured to produce 1000 unique images per epoch, early stopping after 175 epochs in the first stage and 81 epochs in the second stage. 

Similar to the above, we adopted a two-stage approach for training each CNN, training first on simulations with harder examples and then honing in on cases that are closer to the dataset. The generator was configured to present the model with 1000 unique samples per epoch, with batch sizes of 4 for the CR CNNs, 16 for the pupil CNN at 500 Hz, and 8 for the pupil-CNN at 1000 Hz. The batch size was further reduced to 4 for both pupil CNNs during the second stage of training. Additionally, a set of pre-generated synthetic images was used for validation, with a validation set size of 300 for the CR CNNs and 600 for the pupil CNNs. We employed the mean squared error (MSE) loss function for the CR CNNs and the mean absolute error (MAE) loss function for the pupil CNNs, and to assess model performance. To train the models, we used the Adam\cite{kingma2014adam} optimizer for the CR CNNs and the pupil CNN at 1000 Hz, while AdamW was used for the pupil CNN at 500 Hz. In the first stage, the initial learning rate was set to $1e^{-4}$, which was subsequently decreased to $1e^{-6}$ in the second stage. An exponential decay scheduler was used for the learning rate in all training regimes.

The CR CNNs at 500 Hz and 1000 Hz were trained for a maximum of 700 epochs for the first and second stages, incorporating an early stopping mechanism to prevent overfitting. The first stage of the CR CNN at 500 Hz converged after 286 epochs while the second stage required 555 epochs. The 1000 Hz model reached convergence in 167 epochs for the first stage and 307 epochs for the second stage.


In the first stage, the pupil CNNs are allowed to train up to 500 epochs with a patience of 20. In the second stage, the 500 Hz model is trained for up to 40 epochs with a patience of 5 while the 1000 Hz model is trained for up to 100 epochs with a patience of 10. The 500 Hz model reached convergence after 99 epochs in the first stage and 25 epochs in the second stage. The 1000 Hz model achieved convergence after 88 epochs in the first stage and 36 epochs in the second stage. 

In the second stage, the first convolutional layer of each model is frozen and we used an iterative approach to determine which additional layers to freeze, if any. We chose to freeze the first convolutional layer for the pupil CNNs and the 1000Hz CR CNN, and the first two layers of the CR CNN at 500Hz. 

\subsection*{High-Resolution Eye-Tracking Data Collection}

High-resolution eye images were recorded from the first, third and last author of the current paper and one further experienced participant with the FLEX setup \cite{nystrom2023amplitude,hooge2021pupil}. Eye movement data were simultaneously recorded with the EyeLink 1000 Plus (SR Research Ltd., Ottawa, Canada). The setup is shown in Figure \ref{fig:flex}. The EyeLink illuminator was used to illuminate the eye and create the corneal reflection used by both the EyeLink and the FLEX setups. The FLEX setup used a Basler ace acA2500-60um camera equipped with a 50-mm lens (AZURE-5022ML12M) and a near-IR long pass filter (MIDOPT LP715-37.5) that was positioned 50 cm from the participant's eyes.

Two datasets were collected using the same participants and tasks: the FLEX 1) acquired images at 1000 Hz and 2) acquired images at 500 Hz. Camera and illuminator settings for the two data sets were as follows:
\begin{enumerate}
    \item \textit{1000 Hz}. 8-bit images were captured at 672 x 340 pixels, with camera exposure set to \qty{882}{\micro\second} and gain to 12 dB. EyeLink illuminator power was 100\%.
    \item \textit{500 Hz}. 8-bit images were captured at 896 x 600 pixels, with camera exposure set to \qty{1876}{\micro\second} and gain to 10 dB. EyeLink illuminator power was 75\%.
\end{enumerate}
Videos were captured with custom software that streamed the recorded frames to mp4 files using libavcodec (FFMpeg) version 5.1.1 and the libx264 h.264 encoder (preset: veryfast, crf: 17, pixel format: gray).

For both datasets, simultaneous binocular eye movement recordings were performed at 1000 Hz with an EyeLink 1000 Plus (host software 5.12) in desktop setup using the center-of-mass pupil tracking mode. The EyeLink camera sensor was located 56 cm away from the participant’s eyes. To synchronize the acquisition of eye images from the FLEX with eye movement data from the EyeLink, TTL triggers were sent to the EyeLink Host computer at the onset and offset of each FLEX image recording trial. 
The recordings took place in a dark room with no windows.

Several tasks were shown on an Asus VG248QE monitor at 60 Hz (viewing distance 79 cm). Participants performed the following tasks while stabilized on a chin- and forehead rest:
\begin{enumerate}
    \item Nine 1-second fixations in random order on a 3×3 grid of fixation points positioned at $h = \{-7, 0, 7\}$ deg and $v = \{-5, 0, 5\}$ deg.
    \item One 30-second fixation on a point positioned at $h = 0$ deg and $v = 0$ deg while the background luminance alternated between black and white at a cycle time of \qty{3}{\second}.
    \item Three 30-second fixations on points positioned at $h = \{-3.5, 0, 3.5\}$ deg and $v = 0$ deg on a middle grey background, with each position repeated 2 times.
    \item Five rightward step-ramp pursuit stimuli from $h=-10$ deg to $h=10$ deg at a speed of \qty{2}{\degree\per\second} following a \qty{200}{\milli\second} leftward step.
    \item Fixations on a dot that was presented for 1 second at positions $(x, 0), x \in \{-7,\allowbreak -3.5,\allowbreak 0,\allowbreak 3.5,\allowbreak 7\}$ deg, with each position repeated 6 times.
    \item Fifteen fixations in random order on a dot that was presented for 1.5 seconds at positions $h = \{-7,\allowbreak -3.5,\allowbreak 0,\allowbreak 3.5,\allowbreak 7\}$ deg and $v = \{-5,\allowbreak 0,\allowbreak 5\}$ deg, with each position repeated 6 times.
\end{enumerate}
The fixation point consisted of a blue disk (\qty{1.2}{\degree} diameter) with a red point (\qty{0.2}{\degree} diameter) placed on its center.

The total recording time for each participant was approximately 8.5 min, resulting in a database containing approximately 437500 FLEX eye images per participant at 1000 Hz and 219300 images at 500 Hz, along with the EyeLink data.

\subsubsection*{High resolution eye image analysis}

Image analysis was performed frame-wise and adapted from \cite{nystrom2023amplitude} and \cite{byrne2023CRCNN}. In a first stage, pupil and CR centers were localized using the thresholding method. Briefly, fixed thresholds and analysis ROIs were manually set per participant to identify the pupil and CR in the images. The analysis was performed at different pupil and CR thresholds for each participant, and the threshold that resulted in the best precision pupil and CR signals were used. Using these thresholds the images were binarized and after morphological operations to fill holes, the pupil and CR were selected based on shape and size criteria. The center of mass of these binary blobs were then computed; these will be referred to as the pupil and CR centers localized using the thresholding method. For the pupil an ellipse was furthermore fit to the binary pupil blob.

In a second stage, for both the pupil and the CR, 180×180 pixel cutouts centered on the pupil and CR center locations identified by the thresholding method were made. To determine the CR center with the CNN, as was done in \cite{byrne2023CRCNN}, a black circular mask with a 48-pixel radius was applied to the cutout before feeding it into the CR CNN. Similarly, before providing the pupil cutout to the pupil CNN, a middle gray elliptical mask was applied to the cutout that was 1.4 times larger than the pupil ellipse determined in stage 1. RMS-S2S precision \cite{HoNyMu2012,niehorster2020impact,niehorster2020characterizing} of the pupil and CR center locations estimated by both the thresholding and CNN methods was computed in a \qty{200}{\milli\second} window moved over the signals, after which each trial and signal's median RMS values were determined \cite{hooge2022robust,hooge2018human,niehorster2020glassesviewer}.

We computed calibrated gaze signals by subtracting the CR center location from the pupil center location and calibrating the resulting vector with data from the 3x3 grid of fixation points from the first task. We used second-order polynomials in $x$ and $y$ with first-order interactions to calibrate these P--CR signals \cite{cerrolaza2012error,stampe1993heuristic}. To examine the quality of the resulting calibrated gaze data, we computed accuracy as the offset between the estimated gaze location and the target location for the gaze data from task 6, which involved repeated fixations on 15 targets. We determined the RMS-S2S precision of the calibrated gaze signals for all recorded trials in the same way as for the pupil and CR center signals, and computed the standard deviation of the signals using the same sliding window technique.

\subsubsection*{Center of mass calculations}

In order to determine the center of mass or centroid of a feature, specifically the pupil within an image, we employed the following equations~\cite{shortis1994comparison}:
\begin{equation}
CoM_{x} = \sum_{j=1}^{m} \sum_{i=1}^{n} j \cdot  I(i,j) / \sum_{j=1}^{m} \sum_{i=1}^{n} I(i,j) 
\end{equation}
\begin{equation}
CoM_{y} = \sum_{j=1}^{m} \sum_{i=1}^{n} i \cdot  I(i,j) / \sum_{j=1}^{m} \sum_{i=1}^{n} I(i,j)
\end{equation}
where $I(i, j)$ represents the pixel intensity value at row $i$ and column $j$ in an image $I$, and $(m, n)$ denote the dimensions of the image. 

These equations were also used to determine the pupil center location from the annotations provided in the OpenEDS 2019 dataset~\cite{garbin2020dataset} and OpenEDS 2020 dataset~\cite{palmero2021openeds2020}.

\section*{Code and Data Availability}

We have made the code used to generate LEyes simulations and the models employed in our experiments are available on our GitHub repository at \url{https://github.com/dcnieho/Byrneetal_LEyes}. We do not have permission to share the high resolution eye videos we have collected.

\printbibliography

@inproceedings{garbin2020dataset,
  title={Dataset for eye tracking on a virtual reality platform},
  author={Garbin, Stephan Joachim and Komogortsev, Oleg and Cavin, Robert and Hughes, Gregory and Shen, Yiru and Schuetz, Immo and Talathi, Sachin S},
  booktitle={ACM Symposium on Eye Tracking Research and Applications},
  pages={1--10},
  year={2020}
}

@inproceedings{shortis1994comparison,
  title={Comparison of some techniques for the subpixel location of discrete target images},
  author={Shortis, Mark R and Clarke, Timothy A and Short, Tim},
  booktitle={Videometrics III},
  volume={2350},
  pages={239--250},
  year={1994},
  organization={International Society for Optics and Photonics}
}

@article{valliappan2020accelerating,
  title={Accelerating eye movement research via accurate and affordable smartphone eye tracking},
  author={Valliappan, Nachiappan and Dai, Na and Steinberg, Ethan and He, Junfeng and Rogers, Kantwon and Ramachandran, Venky and Xu, Pingmei and Shojaeizadeh, Mina and Guo, Li and Kohlhoff, Kai and others},
  journal={Nature communications},
  volume={11},
  number={1},
  pages={4553},
  year={2020},
  publisher={Nature Publishing Group UK London}
}

@inproceedings{krafka2016eye,
  title={Eye tracking for everyone},
  author={Krafka, Kyle and Khosla, Aditya and Kellnhofer, Petr and Kannan, Harini and Bhandarkar, Suchendra and Matusik, Wojciech and Torralba, Antonio},
  booktitle={Proceedings of the IEEE conference on computer vision and pattern recognition},
  pages={2176--2184},
  year={2016}
}

@inproceedings{niehorster2020towards,
  title={Towards eye tracking as a support tool for pilot training and assessment},
  author={Niehorster, Diederick C and Hildebrandt, Michael and Smoker, Anthony and Jarodzka, Halszka and Dahlstr{\"o}hm, Nicklas},
  booktitle={Eye-tracking in aviation. Proceedings of the 1st international workshop (ETAVI 2020)},
  pages={17--28},
  year={2020},
  organization={ISAE-SUPAERO, Universit{\'e} de Toulouse}
}

@article{byrne2023predicting,
  title={Predicting choice behaviour in economic games using gaze data encoded as scanpath images},
  author={Byrne, Sean Anthony and Reynolds, Adam Peter Frederick and Biliotti, Carolina and Bargagli-Stoffi, Falco J and Polonio, Luca and Riccaboni, Massimo},
  journal={Scientific Reports},
  volume={13},
  number={1},
  pages={4722},
  year={2023},
  publisher={Nature Publishing Group UK London}
}

@misc{byrne2023CRCNN,
      title={Precise localization of corneal reflections in eye images using deep learning trained on synthetic data}, 
      author={Sean Anthony Byrne and Marcus Nyström and Virmarie Maquiling and Enkelejda Kasneci and Diederick C. Niehorster},
      year={2023},
      eprint={2304.05673},
      archivePrefix={arXiv},
      primaryClass={cs.CV}
}

@article{walton2021beyond,
  title={Beyond blur: Real-time ventral metamers for foveated rendering},
  author={Walton, David R and Dos Anjos, Rafael Kuffner and Friston, Sebastian and Swapp, David and Ak{\c{s}}it, Kaan and Steed, Anthony and Ritschel, Tobias},
  journal={ACM Transactions on Graphics},
  volume={40},
  number={4},
  pages={1--14},
  year={2021},
  publisher={Association for Computing Machinery (ACM)}
}

@inproceedings{yassien2020design,
  title={A design space for social presence in VR},
  author={Yassien, Amal and ElAgroudy, Passant and Makled, Elhassan and Abdennadher, Slim},
  booktitle={Proceedings of the 11th Nordic Conference on Human-Computer Interaction: Shaping Experiences, Shaping Society},
  pages={1--12},
  year={2020}
}

@inproceedings{meena2017multimodal,
  title={A multimodal interface to resolve the Midas-Touch problem in gaze controlled wheelchair},
  author={Meena, Yogesh Kumar and Cecotti, Hubert and Wong-Lin, KongFatt and Prasad, Girijesh},
  booktitle={2017 39th Annual International Conference of the IEEE Engineering in Medicine and Biology Society (EMBC)},
  pages={905--908},
  year={2017},
  organization={IEEE}
}

@article{niehorster2019searching,
  title={Searching with and against each other: Spatiotemporal coordination of visual search behavior in collaborative and competitive settings},
  author={Niehorster, Diederick C and Cornelissen, Tim and Holmqvist, Kenneth and Hooge, Ignace T C},
  journal={Attention, Perception, \& Psychophysics},
  volume={81},
  number={3},
  pages={666--683},
  year={2019},
  doi={10.3758/s13414-018-01640-0},
  publisher={Springer}
}

@inproceedings{santini2019grip,
author = {Santini, Thiago and Niehorster, Diederick C. and Kasneci, Enkelejda},
title = {Get a Grip: Slippage-Robust and Glint-Free Gaze Estimation for Real-Time Pervasive Head-Mounted Eye Tracking},
year = {2019},
isbn = {9781450367097},
publisher = {Association for Computing Machinery},
address = {New York, NY, USA},
url = {https://doi.org/10.1145/3314111.3319835},
doi = {10.1145/3314111.3319835},
booktitle = {Proceedings of the 11th ACM Symposium on Eye Tracking Research \& Applications},
articleno = {17},
numpages = {10},
location = {Denver, Colorado},
series = {ETRA '19}
}

@article{stampe1993heuristic,
  title={Heuristic filtering and reliable calibration methods for video-based pupil-tracking systems},
  author={Stampe, Dave M},
  journal={Behavior Research Methods, Instruments, \& Computers},
  volume={25},
  number={2},
  pages={137--142},
  year={1993},
  publisher={Springer}
}

@inproceedings{cerrolaza2012error,
  title={Error characterization and compensation in eye tracking systems},
  author={Cerrolaza, Juan J and Villanueva, Arantxa and Villanueva, Maria and Cabeza, Rafael},
  booktitle={Proceedings of the symposium on eye tracking research and applications},
  pages={205--208},
  year={2012}
}

@inproceedings{HoNyMu2012,
	Author = {Holmqvist, Kenneth and Nystr{\"o}m, Marcus and Mulvey, Fiona},
	Booktitle = {Proceedings of the Symposium on Eye Tracking Research and Applications},
	Organization = {ACM},
	Pages = {45--52},
	Title = {Eye tracker data quality: What it is and how to measure it},
	Year = {2012}
}

@article{niehorster2020characterizing,
  title={Characterizing gaze position signals and synthesizing noise during fixations in eye-tracking data},
  author={Niehorster, Diederick C and Zemblys, Raimondas and Beelders, Tanya and Holmqvist, Kenneth},
  journal={Behavior Research Methods},
  year={2020},
  volume={52},
  number={6},
  pages={2515--2534},
  doi = {10.3758/s13428-020-01400-9},
  publisher={Springer}
}

@article{niehorster2020impact,
  title={The impact of slippage on the data quality of head-worn eye trackers},
  author={Niehorster, Diederick C and Santini, Thiago and Hessels, Roy S and Hooge, Ignace T C and Kasneci, Enkelejda and Nystr{\"o}m, Marcus},
  journal={Behavior Research Methods},
  volume={52},
  number={3},
  pages={1140--1160},
  year={2020},
  doi = {10.3758/s13428-019-01307-0},
  publisher={Springer}
}

@article{niehorster2020glassesviewer,
  title={GlassesViewer: Open-source software for viewing and analyzing data from the Tobii Pro Glasses 2 eye tracker},
  author={Niehorster, Diederick C and Hessels, Roy S and Benjamins, Jeroen S},
  journal={Behavior Research Methods},
  volume={52},
  number={3},
  pages={1244--1253},
  year={2020},
  publisher={Springer}
}

@article{hooge2018human,
  title={Is human classification by experienced untrained observers a gold standard in fixation detection?},
  author={Hooge, Ignace T C and Niehorster, Diederick C and Nystr{\"o}m, Marcus and Andersson, Richard and Hessels, Roy S},
  journal={Behavior Research Methods},
  volume={50},
  number={5},
  pages={1864--1881},
  year={2018},
  publisher={Springer}
}

@article{hooge2022robust,
  title={How robust are wearable eye trackers to slow and fast head and body movements?},
  author={Hooge, Ignace T C and Niehorster, Diederick C and Hessels, Roy S and Benjamins, Jeroen S and Nystr{\"o}m, Marcus},
  journal={Behavior Research Methods},
  pages={1--15},
  year={2022},
  publisher={Springer}
}

@article{pierce2016eye,
  title={Eye tracking reveals abnormal visual preference for geometric images as an early biomarker of an autism spectrum disorder subtype associated with increased symptom severity},
  author={Pierce, Karen and Marinero, Steven and Hazin, Roxana and McKenna, Benjamin and Barnes, Cynthia Carter and Malige, Ajith},
  journal={Biological psychiatry},
  volume={79},
  number={8},
  pages={657--666},
  year={2016},
  publisher={Elsevier}
}

@article{hessels2019eye,
  title={Eye tracking in developmental cognitive neuroscience--The good, the bad and the ugly},
  author={Hessels, Roy S and Hooge, Ignace TC},
  journal={Developmental cognitive neuroscience},
  volume={40},
  pages={100710},
  year={2019},
  publisher={Elsevier}
}

@article{lahey2016power,
  title={The power of eye tracking in economics experiments},
  author={Lahey, Joanna N and Oxley, Douglas},
  journal={American Economic Review},
  volume={106},
  number={5},
  pages={309--313},
  year={2016},
  publisher={American Economic Association 2014 Broadway, Suite 305, Nashville, TN 37203}
}

@article{strohmaier2020eye,
  title={Eye-tracking methodology in mathematics education research: A systematic literature review},
  author={Strohmaier, Anselm R and MacKay, Kelsey J and Obersteiner, Andreas and Reiss, Kristina M},
  journal={Educational Studies in Mathematics},
  volume={104},
  pages={147--200},
  year={2020},
  publisher={Springer}
}

@inproceedings{kim2019nvgaze,
  title={Nvgaze: An anatomically-informed dataset for low-latency, near-eye gaze estimation},
  author={Kim, Joohwan and Stengel, Michael and Majercik, Alexander and De Mello, Shalini and Dunn, David and Laine, Samuli and McGuire, Morgan and Luebke, David},
  booktitle={Proceedings of the 2019 CHI conference on human factors in computing systems},
  pages={1--12},
  year={2019}
}

@inproceedings{nair2020rit,
  title={RIT-Eyes: Rendering of near-eye images for eye-tracking applications},
  author={Nair, Nitinraj and Kothari, Rakshit and Chaudhary, Aayush K and Yang, Zhizhuo and Diaz, Gabriel J and Pelz, Jeff B and Bailey, Reynold J},
  booktitle={ACM Symposium on Applied Perception 2020},
  pages={1--9},
  year={2020}
}

@article{kar2017review,
  title={A review and analysis of eye-gaze estimation systems, algorithms and performance evaluation methods in consumer platforms},
  author={Kar, Anuradha and Corcoran, Peter},
  journal={IEEE Access},
  volume={5},
  pages={16495--16519},
  year={2017},
  publisher={IEEE}
}

@inproceedings{10.1145/3588015.3589197,
author = {Byrne, Sean Anthony and Castner, Nora and Kastrati, Ard and P\l{}omecka, Martyna Beata and Schaefer, William and Kasneci, Enkelejda and Bylinskii, Zoya},
title = {Leveraging Eye Tracking in Digital Classrooms: A Step Towards Multimodal Model for Learning Assistance},
year = {2023},
isbn = {9798400701504},
publisher = {Association for Computing Machinery},
address = {New York, NY, USA},
url = {https://doi.org/10.1145/3588015.3589197},
doi = {10.1145/3588015.3589197},
booktitle = {Proceedings of the 2023 Symposium on Eye Tracking Research and Applications},
articleno = {80},
numpages = {6},
location = {Tubingen, Germany},
series = {ETRA '23}
}

@article{Ellseg_gen,
author = {Kothari, Rakshit S. and Bailey, Reynold J. and Kanan, Christopher and Pelz, Jeff B. and Diaz, Gabriel J.},
title = {EllSeg-Gen, towards Domain Generalization for Head-Mounted Eyetracking},
year = {2022},
issue_date = {May 2022},
publisher = {Association for Computing Machinery},
address = {New York, NY, USA},
volume = {6},
number = {ETRA},
url = {https://doi.org/10.1145/3530880},
doi = {10.1145/3530880},
journal = {Proc. ACM Hum.-Comput. Interact.},
month = {may},
articleno = {139},
numpages = {17}
}

@inproceedings{castner2020deep,
  title={Deep semantic gaze embedding and scanpath comparison for expertise classification during OPT viewing},
  author={Castner, Nora and Kuebler, Thomas C and Scheiter, Katharina and Richter, Juliane and Eder, Th{\'e}r{\'e}se and H{\"u}ttig, Fabian and Keutel, Constanze and Kasneci, Enkelejda},
  booktitle={ACM symposium on eye tracking research and applications},
  pages={1--10},
  year={2020}
}

@article{hooge2021pupil,
  title={The pupil-size artefact (PSA) across time, viewing direction, and different eye trackers},
  author={Hooge, Ignace T C and Niehorster, Diederick C and Hessels, Roy S and Cleveland, Dixon and Nystr{\"o}m, Marcus},
  journal={Behavior Research Methods},
  volume={53},
  number={5},
  pages={1986--2006},
  year={2021},
  publisher={Springer}
}

@article{chetwood2012collaborative,
  title={Collaborative eye tracking: a potential training tool in laparoscopic surgery},
  author={Chetwood, Andrew SA and Kwok, Ka-Wai and Sun, Loi-Wah and Mylonas, George P and Clark, James and Darzi, Ara and Yang, Guang-Zhong},
  journal={Surgical endoscopy},
  volume={26},
  pages={2003--2009},
  year={2012},
  publisher={Springer}
}

@article{tien2014eye,
  title={Eye tracking for skills assessment and training: a systematic review},
  author={Tien, Tony and Pucher, Philip H and Sodergren, Mikael H and Sriskandarajah, Kumuthan and Yang, Guang-Zhong and Darzi, Ara},
  journal={journal of surgical research},
  volume={191},
  number={1},
  pages={169--178},
  year={2014},
  publisher={Elsevier}
}

@article{helgadottir2019digital,
  title={Digital video microscopy enhanced by deep learning},
  author={Helgadottir, Saga and Argun, Aykut and Volpe, Giovanni},
  journal={Optica},
  volume={6},
  number={4},
  pages={506--513},
  year={2019},
  publisher={Optica Publishing Group}
}

@misc{Iakubovskii:2019,
  Author = {Pavel Iakubovskii},
  Title = {Segmentation Models Pytorch},
  Year = {2019},
  Publisher = {GitHub},
  Journal = {GitHub repository},
  Howpublished = {\url{https://github.com/qubvel/segmentation_models.pytorch}}
}

@misc{ronneberger2015unet,
      title={U-Net: Convolutional Networks for Biomedical Image Segmentation}, 
      author={Olaf Ronneberger and Philipp Fischer and Thomas Brox},
      year={2015},
      eprint={1505.04597},
      archivePrefix={arXiv},
      primaryClass={cs.CV}
}

@INPROCEEDINGS{5206848,

  author={Deng, Jia and Dong, Wei and Socher, Richard and Li, Li-Jia and Kai Li and Li Fei-Fei},

  booktitle={2009 IEEE Conference on Computer Vision and Pattern Recognition}, 

  title={ImageNet: A large-scale hierarchical image database}, 

  year={2009},

  volume={},

  number={},

  pages={248-255},

  doi={10.1109/CVPR.2009.5206848}}

@article{koonce2021resnet,
  title={ResNet 34},
  author={Koonce, Brett and Koonce, Brett},
  journal={Convolutional Neural Networks with Swift for Tensorflow: Image Recognition and Dataset Categorization},
  pages={51--61},
  year={2021},
  publisher={Springer}
}

@misc{loshchilov2019decoupled,
      title={Decoupled Weight Decay Regularization}, 
      author={Ilya Loshchilov and Frank Hutter},
      year={2019},
      eprint={1711.05101},
      archivePrefix={arXiv},
      primaryClass={cs.LG}
}

@article{kingma2014adam,
  title={Adam: A method for stochastic optimization},
  author={Kingma, Diederik P and Ba, Jimmy},
  journal={arXiv preprint arXiv:1412.6980},
  year={2014}
}

@article{ruby2020binary,
  title={Binary cross entropy with deep learning technique for image classification},
  author={Ruby, Usha and Yendapalli, Vamsidhar},
  journal={Int. J. Adv. Trends Comput. Sci. Eng},
  volume={9},
  number={10},
  year={2020}
}

@inproceedings{lin2017focal,
  title={Focal loss for dense object detection},
  author={Lin, Tsung-Yi and Goyal, Priya and Girshick, Ross and He, Kaiming and Doll{\'a}r, Piotr},
  booktitle={Proceedings of the IEEE international conference on computer vision},
  pages={2980--2988},
  year={2017}
}

@article{li2019dice,
  title={Dice loss for data-imbalanced NLP tasks},
  author={Li, Xiaoya and Sun, Xiaofei and Meng, Yuxian and Liang, Junjun and Wu, Fei and Li, Jiwei},
  journal={arXiv preprint arXiv:1911.02855},
  year={2019}
}

@article{fuhl2016pupilnet,
  title={Pupilnet: Convolutional neural networks for robust pupil detection},
  author={Fuhl, Wolfgang and Santini, Thiago and Kasneci, Gjergji and Kasneci, Enkelejda},
  journal={arXiv preprint arXiv:1601.04902},
  year={2016}
}

@article{fuhl2017pupilnet,
  title={Pupilnet v2. 0: Convolutional neural networks for cpu based real time robust pupil detection},
  author={Fuhl, Wolfgang and Santini, Thiago and Kasneci, Gjergji and Rosenstiel, Wolfgang and Kasneci, Enkelejda},
  journal={arXiv preprint arXiv:1711.00112},
  year={2017}
}

@misc{fuhl2023pistol,
      title={Pistol: Pupil Invisible Supportive Tool to extract Pupil, Iris, Eye Opening, Eye Movements, Pupil and Iris Gaze Vector, and 2D as well as 3D Gaze}, 
      author={Wolfgang Fuhl and Daniel Weber and Shahram Eivazi},
      year={2023},
      eprint={2201.06799},
      archivePrefix={arXiv},
      primaryClass={cs.CV}
}

@article{bansal2022systematic,
  title={A systematic review on data scarcity problem in deep learning: solution and applications},
  author={Bansal, Ms Aayushi and Sharma, Dr Rewa and Kathuria, Dr Mamta},
  journal={ACM Computing Surveys (CSUR)},
  volume={54},
  number={10s},
  pages={1--29},
  year={2022},
  publisher={ACM New York, NY}
}

@article{kothari2022ellseg,
  title={EllSeg-Gen, towards Domain Generalization for head-mounted eyetracking},
  author={Kothari, Rakshit S and Bailey, Reynold J and Kanan, Christopher and Pelz, Jeff B and Diaz, Gabriel J},
  journal={Proceedings of the ACM on Human-Computer Interaction},
  volume={6},
  number={ETRA},
  pages={1--17},
  year={2022},
  publisher={ACM New York, NY, USA}
}

@InProceedings{Wood_2015_ICCV,
author = {Wood, Erroll and Baltrusaitis, Tadas and Zhang, Xucong and Sugano, Yusuke and Robinson, Peter and Bulling, Andreas},
title = {Rendering of Eyes for Eye-Shape Registration and Gaze Estimation},
booktitle = {Proceedings of the IEEE International Conference on Computer Vision (ICCV)},
month = {December},
year = {2015}
}

@article{santini2018pure,
  title={PuRe: Robust pupil detection for real-time pervasive eye tracking},
  author={Santini, Thiago and Fuhl, Wolfgang and Kasneci, Enkelejda},
  journal={Computer Vision and Image Understanding},
  volume={170},
  pages={40--50},
  year={2018},
  publisher={Elsevier}
}

@article{midtvedt2021quantitative,
  title={Quantitative digital microscopy with deep learning},
  author={Midtvedt, Benjamin and Helgadottir, Saga and Argun, Aykut and Pineda, Jes{\'u}s and Midtvedt, Daniel and Volpe, Giovanni},
  journal={Applied Physics Reviews},
  volume={8},
  number={1},
  pages={011310},
  year={2021},
  publisher={AIP Publishing LLC}
}

@article{yiu2019deepvog,
  title={DeepVOG: Open-source pupil segmentation and gaze estimation in neuroscience using deep learning},
  author={Yiu, Yuk-Hoi and Aboulatta, Moustafa and Raiser, Theresa and Ophey, Leoni and Flanagin, Virginia L and Zu Eulenburg, Peter and Ahmadi, Seyed-Ahmad},
  journal={Journal of neuroscience methods},
  volume={324},
  pages={108307},
  year={2019},
  publisher={Elsevier}
}

@article{10.1145/3591130,
author = {Byrne, Sean Anthony and Maquiling, Virmarie and Reynolds, Adam Peter Frederick and Polonio, Luca and Castner, Nora and Kasneci, Enkelejda},
title = {Exploring the Effects of Scanpath Feature Engineering for Supervised Image Classification Models},
year = {2023},
issue_date = {May 2023},
publisher = {Association for Computing Machinery},
address = {New York, NY, USA},
volume = {7},
number = {ETRA},
url = {https://doi.org/10.1145/3591130},
doi = {10.1145/3591130},
journal = {Proc. ACM Hum.-Comput. Interact.},
month = {may},
articleno = {161},
numpages = {18}
}

@inproceedings{renner2017attention,
  title={Attention guiding techniques using peripheral vision and eye tracking for feedback in augmented-reality-based assistance systems},
  author={Renner, Patrick and Pfeiffer, Thies},
  booktitle={2017 IEEE symposium on 3D user interfaces (3DUI)},
  pages={186--194},
  year={2017},
  organization={IEEE}
}

@article{chaudhary2022temporal,
  title={Temporal RIT-Eyes: From real infrared eye-images to synthetic sequences of gaze behavior},
  author={Chaudhary, Aayush K and Nair, Nitinraj and Bailey, Reynold J and Pelz, Jeff B and Talathi, Sachin S and Diaz, Gabriel J},
  journal={IEEE Transactions on Visualization and Computer Graphics},
  volume={28},
  number={11},
  pages={3948--3958},
  year={2022},
  publisher={IEEE}
}

@inproceedings{chugh2021detection,
  title={Detection and Correspondence Matching of Corneal Reflections for Eye Tracking Using Deep Learning},
  author={Chugh, Soumil and Brousseau, Braiden and Rose, Jonathan and Eizenman, Moshe},
  booktitle={2020 25th International Conference on Pattern Recognition (ICPR)},
  pages={2210--2217},
  year={2021},
  organization={IEEE}
}

@article{gao2023synthetic,
  title={Synthetic data accelerates the development of generalizable learning-based algorithms for X-ray image analysis},
  author={Gao, Cong and Killeen, Benjamin D and Hu, Yicheng and Grupp, Robert B and Taylor, Russell H and Armand, Mehran and Unberath, Mathias},
  journal={Nature Machine Intelligence},
  volume={5},
  number={3},
  pages={294--308},
  year={2023},
  publisher={Nature Publishing Group UK London}
}

@inproceedings{osinski2020simulation,
  title={Simulation-based reinforcement learning for real-world autonomous driving},
  author={Osi{\'n}ski, B{\l}a{\.z}ej and Jakubowski, Adam and Zi{\k{e}}cina, Pawe{\l} and Mi{\l}o{\'s}, Piotr and Galias, Christopher and Homoceanu, Silviu and Michalewski, Henryk},
  booktitle={2020 IEEE international conference on robotics and automation (ICRA)},
  pages={6411--6418},
  year={2020},
  organization={IEEE}
}

@inproceedings{park2018learning,
  title={Learning to find eye region landmarks for remote gaze estimation in unconstrained settings},
  author={Park, Seonwook and Zhang, Xucong and Bulling, Andreas and Hilliges, Otmar},
  booktitle={Proceedings of the 2018 ACM symposium on eye tracking research \& applications},
  pages={1--10},
  year={2018}
}

@article{hooge2016pupil,
  title={The pupil is faster than the corneal reflection (CR): Are video based pupil-CR eye trackers suitable for studying detailed dynamics of eye movements?},
  author={Hooge, Ignace and Holmqvist, Kenneth and Nystr{\"o}m, Marcus},
  journal={Vision research},
  volume={128},
  pages={6--18},
  year={2016},
  publisher={Elsevier}
}

@article{palmero2021openeds2020,
  title={Openeds2020 challenge on gaze tracking for vr: Dataset and results},
  author={Palmero, Cristina and Sharma, Abhishek and Behrendt, Karsten and Krishnakumar, Kapil and Komogortsev, Oleg V and Talathi, Sachin S},
  journal={Sensors},
  volume={21},
  number={14},
  pages={4769},
  year={2021},
  publisher={MDPI}
}

@inproceedings{niu2021real,
  title={Real-Time Localization and Matching of Corneal Reflections for Eye Gaze Estimation via a Lightweight Network},
  author={Niu, Lijinliang and Gu, Zhaopeng and Ye, Juntao and Dong, Qiulei},
  booktitle={The Ninth International Symposium of Chinese CHI},
  pages={33--40},
  year={2021}
}

@article{wang2023eye,
  title={Eye-UNet: A UNet-based network with attention mechanism for low-quality human eye image segmentation},
  author={Wang, Yanxia and Wang, Jingyi and Guo, Ping},
  journal={Signal, Image and Video Processing},
  volume={17},
  number={4},
  pages={1097--1103},
  year={2023},
  publisher={Springer}
}

@article{kothari2020gaze,
  title={Gaze-in-wild: A dataset for studying eye and head coordination in everyday activities},
  author={Kothari, Rakshit and Yang, Zhizhuo and Kanan, Christopher and Bailey, Reynold and Pelz, Jeff B and Diaz, Gabriel J},
  journal={Scientific reports},
  volume={10},
  number={1},
  pages={2539},
  year={2020},
  publisher={Nature Publishing Group UK London}
}

@ARTICLE{9153754,
  author={Akinyelu, Andronicus A. and Blignaut, Pieter},
  journal={IEEE Access}, 
  title={Convolutional Neural Network-Based Methods for Eye Gaze Estimation: A Survey}, 
  year={2020},
  volume={8},
  number={},
  pages={142581-142605},
  doi={10.1109/ACCESS.2020.3013540}}

@inproceedings{maquiling2023virnet,
  title     = {V-ir-Net: A Novel Neural Network for Pupil and Corneal Reflection Detection trained on Simulated Light Distributions},
  author    = {Maquiling, Virmarie and Byrne, Sean Anthony and Nystr{\"o}m, Marcus and Kasneci, Enkelejda and Niehorster, Diederick C.},
  booktitle = {25th International Conference on Mobile Human-Computer Interaction (MobileHCI '23 Companion)},
  year      = {2023},
  address   = {Athens, Greece},
  month     = {September 26--29},
  publisher = {ACM},
  doi       = {10.1145/3565066.3608690},
  isbn      = {978-1-4503-9924-1/23/09}
}

@article{niehorster2021apparent,
  title={Is apparent fixational drift in eye-tracking data due to filters or eyeball rotation?},
  author={Niehorster, Diederick C and Zemblys, Raimondas and Holmqvist, Kenneth},
  journal={Behavior Research Methods},
  volume={53},
  pages={311--324},
  year={2021},
  publisher={Springer}
}

@article{nystrom2023amplitude,
  title={The amplitude of small eye movements can be accurately estimated with video-based eye trackers},
  author={Nystr{\"o}m, Marcus and Niehorster, Diederick C and Andersson, Richard and Hessels, Roy S and Hooge, Ignace TC},
  journal={Behavior Research Methods},
  volume={55},
  number={2},
  pages={657--669},
  year={2023},
  publisher={Springer}
}

@inproceedings{kim2019eye,
  title={Eye semantic segmentation with a lightweight model},
  author={Kim, Soo-Hyung and Lee, Guee-Sang and Yang, Hyung-Jeong and others},
  booktitle={2019 IEEE/CVF International Conference on Computer Vision Workshop (ICCVW)},
  pages={3694--3697},
  year={2019},
  organization={IEEE}
}

\section*{Acknowledgements}

We thank Michael Kirchhof, Yao Rong \& Rakshit Kothari for their valuable suggestions and advice. We would also like to sincerely thank Ignace Hooge for not blinking, ever. We gratefully acknowledge the Lund University Humanities Lab where data were recorded.

\section*{Author contributions statement}

D.N., S.B. and M.N. conceived the experiment(s),  D.N., M.N., V.M. and S.B. conducted the experiment(s), S.B., D.N., V.M. and M.N. analyzed the results.  All authors wrote and reviewed the manuscript. 

\end{document}